\documentclass[10pt,twocolumn,journal,compsoc]{IEEEtran}

\usepackage{fancyhdr, epsfig, epsf, amsthm, amsmath, amssymb, amsfonts, subfigure,color}
\usepackage{threeparttable}
\usepackage[noadjust]{cite}
\usepackage{dsfont}
\usepackage{enumerate}
\usepackage{comment}
\usepackage{multirow,booktabs}
\usepackage{bigstrut}
\usepackage[most]{tcolorbox}
\tcbuselibrary{breakable} 
\usepackage{enumitem}
\usepackage{hyperref}

\def\l{\left}
\def\r{\right}
\def\({\left(}
\def\){\right)}

\def\[{\left[}
\def\]{\right]}

\newcommand{\tred}[1]{{\textcolor{red}{#1}}}

\providecommand{\ip}[1]{\boldsymbol{\langle}#1\boldsymbol{\rangle}}
\def\sm{\small}
\def\nm{\normalsize}
\def\bm{\tred{//}}

\definecolor{codegreen}{rgb}{0,0.6,0}
\definecolor{codegray}{rgb}{0.5,0.5,0.5}
\definecolor{codepurple}{rgb}{0.58,0,0.82}
\definecolor{backcolour}{rgb}{0.95,0.95,0.92}

\lstdefinestyle{mystyle}{
    language=Python,
    backgroundcolor=\color{backcolour},   
    commentstyle=\color{codepurple}\bfseries,
    keywordstyle=\color{codegreen}\bfseries,
    numberstyle=\tiny\color{codegray},
    stringstyle=\color{red},
    basicstyle=\ttfamily\footnotesize, 
    breakatwhitespace=false,         
    breaklines=true,                 
    captionpos=b,                    
    keepspaces=false,                 
    numbers=right,                    
    numbersep=1pt,                  
    showspaces=false,                
    showstringspaces=false,
    showtabs=false,                  
    tabsize=2,
    morekeywords={yield},
    emph ={problem_op, mixing_op},
    emphstyle=\color{blue},
    deletekeywords=[2]{compile},
}

\def\papertitle{Communication-Efficient and Distributed Learning Over Wireless Networks: Principles and Applications}

\IEEEoverridecommandlockouts

\begin{document}
\title{ \fontsize{22}{28}\selectfont  \papertitle}

\author{$^\dagger$Jihong~Park, Sumudu~Samarakoon, Anis~Elgabli, $^\ddagger$Joongheon Kim,\\ Mehdi Bennis, $^\star$Seong-Lyun Kim, and $^{\star\star}$M\'{e}rouane Debbah
\thanks{$^\dagger$J.~Park is with the School of Information Technology, Deakin University, VIC~3220, Geelong, Australia (email: jihong.park@deakin.edu.au).}
\thanks{
S.~Samarakoon, A.~Elgabli, and M.~Bennis are with the Centre for Wireless Communications, University of Oulu, Oulu~90014, Finland (email: \{sumudu.samarakoon, anis.elgabli, mehdi.bennis\}@oulu.fi). }
\thanks{J.~Kim is with the School of Electrical Engineering, Korea University, Seoul 02841, Korea (e-mail: joongheon@korea.ac.kr).
}
\thanks{$^*$S.-L.~Kim is with the School of Electrical and Electronic Engineering, Yonsei University, Seoul~03722, Korea (email: slkim@yonsei.ac.kr)}
\thanks{$^{**}$M.~Debbah is with CentraleSup\'{e}lec, Universit\'{e} Paris-Saclay, Gif-sur-Yvette~91190, France (email: merouane.debbah@centralesupelec.fr).}
}

\maketitle \thispagestyle{empty}

\begin{abstract} 
Machine learning (ML) is a promising enabler for the fifth generation (5G) communication systems and beyond. By imbuing intelligence into the network edge, edge nodes can proactively carry out decision-making, and thereby react to local environmental changes and disturbances while experiencing zero communication latency. To achieve this goal, it is essential to cater for high ML inference accuracy at scale under time-varying channel and network dynamics, by continuously exchanging fresh data and ML model updates in a distributed way. Taming this new kind of data traffic boils down to improving the communication efficiency of distributed learning by optimizing communication payload types, transmission techniques, and scheduling, as well as ML architectures, algorithms, and data processing methods. To this end, this article aims to provide a holistic overview of relevant communication and ML principles, and thereby present communication-efficient and distributed learning frameworks with selected use cases.
\end{abstract}




	
	
	




\section{Significance and Motivation}\label{sec:significance}

The pursuit of extremely stringent latency and reliability guarantees is essential in the fifth generation (5G) communication system and beyond \cite{MehdiURLLC:18,park2020extreme}. In a wirelessly automated factory, the remote control of assembly robots should provision the same level of target latency and reliability offered by existing wired factory systems. 
To this end, for instance, control packets should be delivered within 1\,ms with 99.99999\% reliability \cite{wp5d2017minimum,pokhrel2020towards,latva2019key}. 
%
%
In the emerging non-terrestrial communication enabled by a massive constellation of low-orbit satellites \cite{alenSpace,Starlink,Kuniper,Oneweb,lee2020integrating}, the orbiting speed is over 8\,km per second, under which a single emergency control packet loss may incur collisions with other satellites and space debris. 
%
%
Unfortunately, traditional methods postulate known channel and network topological models while focusing primarily on maximizing data rates. Such model-based and best-effort solutions are far from enough to meet the challenging latency and reliability requirements under limited radio resources and randomness on wireless channels and network topologies in practice. 

Realizing the aforementioned pressing concern has recently sparked huge attention to the introduction of machine learning (ML) based approaches into communication system designs \cite{Chen2017MachineLF,park2018wireless,Dorner:18,WangHan:18}. 
By leveraging ML at the network edge, each edge node can proactively carry out decision-making based on its local predictions, thereby experiencing zero latency \cite{park2020extreme,tinyML}.
Furthermore, real data observations construct these ML models that directly reflect the environment in reality without modeling artifacts. In these respects, one may misapprehend that communication becomes less important in 5G and beyond where everything is locally predictable. The answer is the opposite, as accurate ML prediction cannot be achieved and sustained without communication.

\begin{figure*}
    \centering
    \includegraphics[width=\textwidth]{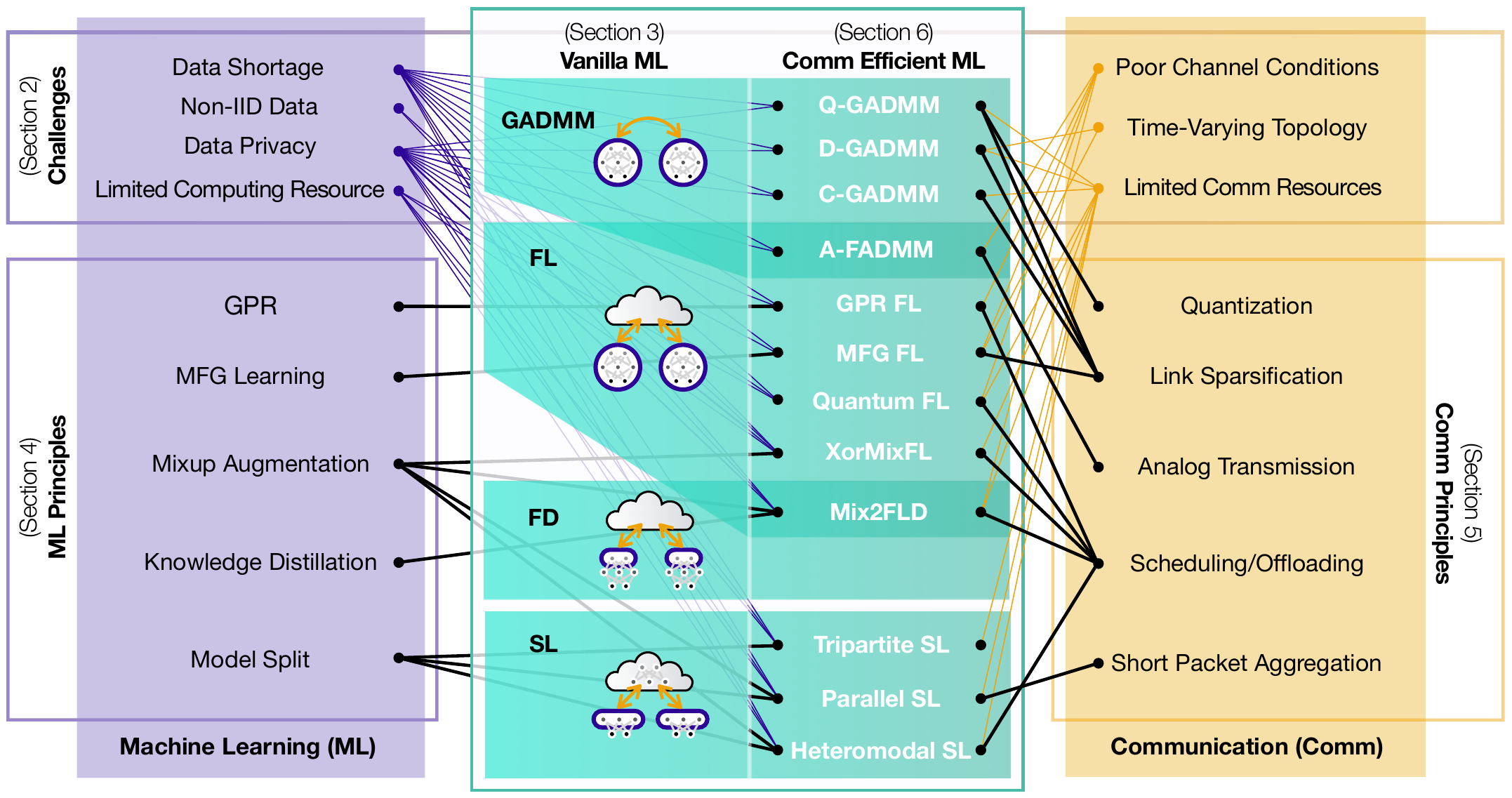}
    \caption{An overview of this article that aims to present communication-efficient and distributed learning frameworks (Sec. \ref{sec:advanced}) by applying advanved machine learning (ML) and communication principles (Sec. Sec.~\ref{sec:principles_comm} and \ref{sec:principles_ml}) to baseline distributed learning algorithms (Sec. \ref{sec:vanilla}) for addressing key ML and communication challenges (Sec. \ref{sec:challenges}).}
    \label{fig:intro_overview}
\end{figure*}

In particular, observing dispersed data is prerequisite to train and run data-driven ML models with high prediction or inference accuracy. The observation does not have to be done by directly collecting the raw data from edge nodes, which may violate their data privacy. Alternatively, by leveraging federated learning (FL), it is possible to exchange ML model parameters that reflect the data observed by each ML model without revealing raw data \cite{pap:jakub16}.
Similarly, one can exchange ML model outputs \cite{OnlineKD,Ahn:PIMRC20} or hidden activations \cite{Vepakomma:2018:Splita} for higher communication efficiency while preserving data privacy.
Such a communication is not a one-time event, since a trained ML model can easily be outdated and should thus be continually re-trained under time-varying data distributions and environments. As a consequence, ML will not only be a key enabler of future communication systems, but also be one major source of data traffic, which warrants taming the new kind of traffic generated by distributed learning. Furthermore, communication environments have a considerable impact on the performance of ML. Indeed, temporal network topology variations and uplink-downlink channel asymmetry determine learning stragglers. 
In analog transmissions, channel fluctuations directly distort communicating information \cite{haykin1994introduction}, affecting the ML accuracy and data privacy. 
This mandates to co-design distributed ML and communication operations.



Spurred by the aforementioned motivations, this article aims to present communication-efficient and distributed learning frameworks built upon jointly optimizing the types of communication payloads, transmissions, and scheduling as well as ML architectures and algorithms under wireless channel dynamics and network topology variations. To reach the overarching goal, as visualized in Fig.~\ref{fig:intro_overview}, this article is structured as follows. 
In Sec.~\ref{sec:challenges}, major technical challenges are summarized. 
In Sec.~\ref{sec:vanilla}, existing communication-efficient and distributed learning frameworks and their limitations are briefly reviewed. 
To improve these vanilla distributed learning frameworks, several ML and communication design principles are discussed in Sec.~\ref{sec:principles_comm} and \ref{sec:principles_ml}. 
Finally, selected applications of such principles and their effectiveness are elaborated in Sec.~\ref{sec:advanced}, followed by concluding remarks in~Sec.~\ref{sec:conclusion}.









\section{Key Challenges}\label{sec:challenges}

Towards understanding the underlying black-box operations of ML, centralized ML architectures have been the prime focus in the theoretical studies.
With the paradigm shift from cloud-centric to on-device ML, above theoretical analysis cannot be readily applicable to investigate the current distributed ML architecture.
In this view, we identify several key challenges that needs to be addressed in designing distributed learning over wireless networks as discussed next.

\vspace{5pt}\noindent\textbf{Data Shortage.}\quad
%
One of the main downsides of shifting from cloud to device in data-driven ML is the limited access to sufficiently large datasets.
Devices with low exposure and storage may not be able to accumulate rich datasets, in which training on-devices may lose generalization and are susceptible to unseen data \cite{mohri2018foundations}. 
Data acquisition within the device could results in higher end-to-end training latencies and/or outdated ML models in the presence of dynamic data.
To overcome the data shortage, robust and collaborative ML designs need to be investigated.

\vspace{5pt}\noindent\textbf{Non-IID Data.}\quad
%
User-generated data could be highly personalized (e.g. different angles and frame rates of surveilance cameras,), imbalanced (e.g., some labels corresponding to extreme events have much fewer samples than other labels), and multimodal (e.g., temperature and humidity sensors for weather prediction), all of which are described as non-independent and identically distributed (IID) data. Under non-IID data, it is common that the accuracy and convergence speed of distributed learning are significantly degraded \cite{Zhao2018,frid2018gan,bowles2018gan,hsu2017unsupervised,nishizaki2017data}. Furthermore, majority of the analytical frameworks are devised on the training with IID data and cannot be easily extended towards distributed learning over non-IID data \cite{kairouz2019advances}.
Hence, deriving convergence characteristic and reliability and robustness guarantees with on-device learning is a daunting task.

\vspace{5pt}\noindent\textbf{Data Privacy.}\quad
%
Data owned by the devices may contain privacy-sensitive information, and thus, exchanging ML model parameters instead data is widely used in distributed learning.
Yet, the exposed model parameters could be reversely traced, in which, privacy is only partially preserved \cite{fredrikson2015model}.
To further enhance privacy, adopting extra coding, introducing noise to shared parameters, and exchanging redundant information are some viable solutions.
However, each of above solutions introduce additional challenges (e.g., increased processing delays with extra coding, loss of inference accuracy due to the excess noise, and extra communication delays with redundant information).
%

\vspace{5pt}\noindent\textbf{Computing Resource Limitation.}\quad
Training and operating ML models require huge computation processor energy, memory, and high-speed inter-processor communication links, which is commonly not available at battery-limited small edge devices.
%
%
Hence, deep learning computations are often carried out at a cloud server using high-performance computing (HPC) resources~\cite{icdcs18} consisting of graphics processing units (GPUs), each of which is equipped with thousands of core processing units (e.g., NVIDIA GTX 2080 Ti has 4\,352 CUDA cores and 544 tensor processing units \cite{TECHRADAR}). 
This cannot be expected to shift to the network edge without simplifying their complexities. 
In addition, due to the limited energy and memory/storage at edge devices, processing low-complex small models and tasks is of the utmost importance. 
In this view, designs of energy-efficient \emph{low-precision} ML and \emph{binary neural networks} (NNs) need to be considered with distributed learning \cite{ibmWanglowprec,intelRodriguezlowprec,murovivc2019massively}.

\vspace{5pt}\noindent\textbf{Communication Resource Limitation.}\quad
%
Relying on the limited wireless resources that are shared among multitude of devices and services and thus, susceptible to high interference and intermittent connectivity can restrain the distributed learning performance and speed \cite{park2018wireless}. 
Mobile service operators are restricted to limited frequency bands as well as bandwidths, in which the difficulties on ensuring reliable and low-latency connectivity for training devices grow exponentially as the network scales.
Introducing more bandwidth to the network via the usage of high frequency bands, e.g., millimeter waves (mmWaves), cannot simply resolve the lack of wireless resources due to their inherited unreliable nature of channel conditions (propagation losses, blockages, and fading) \cite{wang2018millimeter}.
While increasing/optimizing transmit power and adopting encoding-decoding techniques can be beneficial in terms of enhancing reliable connectivity, training devices may not be able to exploit them with the limited power availability \cite{lun2008coding}.
Therefore, communication resource management is a key aspect on realizing distributed learning.





\vspace{5pt}\noindent\textbf{Poor Channel Conditions.}\quad
%
Distributed learning over large number of devices collaborating one another relies on the inter-device communication over wireless links.
Under wireless channel dynamics, communication among devices is likely to be affected by poor channel conditions and the transmission noise yielding increased training latencies and losses in both training and inference accuracy \cite{park2018wireless}. 
With the limited wireless resources, it is crucial to adopt existing communication techniques (e.g., scheduling, coding, quantizing, relaying, interference managing, millimeter wave communication etc.)  and to extend them considering the aspects of distributed learning (e.g., guarantees on training latency, accuracy, reliability, and robustness).

\vspace{5pt}\noindent\textbf{Time-Varying Network Topology.}\quad
%
Mobility is inherited in devices, in which, distributed learning needs to cope with dynamic network topologies.
With time-varying networks, learning agents are affected by loss of connectivity, inconsistent and asynchronous collaboration, frequent model mismatches, and tendency of having outdated data and models \cite{mohri2018foundations}.
Developing distributed training mechanisms and analyzing them over above dynamics is extremely difficult. 
Resorting to predictive/proactive techniques and recasting the interactions among many agents to simplified statistical models are essential for learning over dynamic wireless network topologies.

\section{Related Distributed Learning Methods}\label{sec:vanilla}


Distributed ML algorithms are briefly categorized into the methods exchanging model parameters, model outputs, and hidden activations, with or without the aid of a parameter server. In this section, we introduce representative distributed ML methods, followed by identifying the limitations of these vanilla approaches, calling for applying new key principles and developing advanced ML frameworks to be elaborated in the next sections.

\subsection{Federated Learning (FL)}\label{sec:method_FL}

FL is a distributed training framework, which has been successfully adopted for Google's predictive keyboards \cite{yang2018applied} and many other use cases in the areas of healthcare, intelligent transportation, and industrial automation \cite{samarakoon2019distributed,kairouz2019advances}. In essence, FL is designed to periodically upload workers' model parameters (e.g., NN weights and/or gradients) during local training to a parameter server that performs model averaging and broadcasts the resultant global model to all workers~\cite{konevcny2015federated}. Here, avoiding raw data exchanges preserves data privacy, while adjusting the uploading period improves communication efficiency. 

Recent studies have investigated different training aspects including personalization (i.e., multi-task learning) \cite{jnl:smith17}, robustness guarantees \cite{ARM18,sattler2019robust}, and training over dynamic topologies \cite{lalitha2019peer}. One critical issue of FL is that its communication overhead is proportional to the number of model parameters. Consequently, FL struggles with supporting deep NNs over capacity-limited wireless channels. 






\subsection{Group ADMM (GADMM)}\label{sec:method_gadmm}

The parameter server in FL cannot be connected with faraway workers. Furthermore, it is vulnerable to a single point of attack or failure \cite{KimCL:19}. In this regard, leveraging the alternating direction method of multipliers (ADMM) method, group ADMM (GADMM) aims to enable distributed learning without any central entity while communicating only with neighboring workers \cite{elgabli2020gadmm}. To this end, GADMM divides the workers into {\it head} and {\it tail} groups. Each worker from head or tail group exchanges variables with only two workers from the tail/head group forming a chain. At each iteration, every head worker first updates its primal variable (i.e., models) in parallel by minimizing the augmented Lagrangian function defined in ADMM, while utilizing its two neighboring 
tail workers' models in the previous iteration. Once head workers update their models, each worker transmits its updated model to its two neighbors from the tail group. Then, following the same way, every tail worker updates its model by utilizing its two neighboring head workers' models received in the current iteration. Finally, the dual variables are updated locally at each worker.

With GADMM, at every communication round, only half of the workers are competing for the limited bandwidth. Moreover, by limiting the communication only to the two neighboring workers, the communication energy can significantly be reduced. Nonetheless, GADMM relies on model parameter exchanges as in FL whose communication payload size increases with the number of parameters, limiting the scalability of GADMM particularly under deep NNs.

\subsection{Federated Distillation (FD)}\label{sec:method_FD}
Modern deep NN architectures often have a large number of model parameters. For instance, GPT-3 model is a state-of-the-art NN architecture for natural language processing (NLP) tasks, and has $175$ billion parameters corresponding to over $350$\,GB \cite{brown2020language}. Exchanging the sheer amount of deep NN model parameter is costly, hindering frequent communications particularly under limited wireless resources. Alternatively, FD only exchanges the models' outputs whose dimensions are much smaller than the model sizes (e.g., 10 classes in the MNIST dataset). To illustrate, in a classification task, each worker runs local iterations while storing the average model output (i.e., logit) per class. At a regular interval, these local average outputs are uploaded to the parameter server aggregating and averaging the local average output across workers per class. The resultant global average outputs are downloaded by each worker. Finally, to transfer the downloaded global knowledge into local models, each worker runs local iterations with its own loss function in addition to a regularizer measuring the gap between its own prediction output of a training sample and the global average output for the given class of the sample. Such a regularization method is called knowledge distillation (KD) that is to be detailed in Sec. \ref{sec:principle_KD}. 


FD is not limited to simple classification tasks under a perfectly controlled environment. In \cite{Han:Intellisys20}, FD is extended to an reinforcement learning (RL) application by replacing the aforementioned pre-class averaging step of FD with an averaging operations across neighboring states for an RL task. In \cite{Oh20:CL,Ahn:ICASSP20,Ahn:PIMRC20}, FD is implemented in a wireless fading channel, demonstrating comparable accuracy under channel fluctuations and outages with much less payload sizes compared to FL. Nonetheless, FD is more vulnerable to the problem of non-IID data distributions. Even if a worker obtains the global average outputs for all classes, when the worker lacks the samples in a specific target class, the global knowledge is rarely transferred into the local model of the~worker.

\subsection{Split Learning (SL)} \label{sec:method_SL}

A large-sized deep NN cannot be fit into edge devices' small memory. Split learning (SL) resolves this problem by dividing a single NN into multiple segments and distributing the lower segments across multiple workers storing raw data \cite{Vepakomma:2018:Splita,Vepakomma:2019}. By connecting the lower segments with a shared upper segment stored at a parameter server, each device uploads its NN activations of the cut-layer (i.e., lower segment's last layer) to the server calculating the loss values, and downloads the gradients to update its lower segment. As done in FL, FD, and GADMM, SL also hides raw data, preserving data privacy. For this reason, SL has recently been adopted in medical applications wherein dispersed private health records should be exploited without sharing raw data \cite{Vepakomma:2018:Splita,Vepakomma:2019} \cite{jeon2019privacy}. SL has also been known for its robustness against non-IID data distributions, and applied for fusing heterogeneous vision and radio-frequency (RF) modalities to predict millimeter-wave channels \cite{koda2019one,koda2020commun,koda2020distributed}. 

While effective in terms of accuracy, the communication efficiency of SL is still questionable. As opposed to FL, FD, and GADMM that periodically exchange model updates, SL requires to exchange instantaneous model updates in feed-forward and backward propagations. For some applications, SL yields less communication overhead compared to the aforementioned periodic-update benchmark schemes by achieving much faster convergence \cite{singh2019detailed}, which may not always be feasible under different tasks and datasets. Furthermore, the communication cost of SL depends on the NN architecture and how to cut its NN layers, calling for more investiation on co-desigining its communicataion and NN architectures.


\subsection{Multi-Agent Reinforcement Learning (MARL)} \label{sec:method_MARL}

Thus far we have implicitly considered that the datasets are fixed and independent across different workers. This isolated and stationary dataset assumption is not feasible when each worker interacts with other workers in a common environment, while making decisions based on its own observation of the environment. Multi-agent reinforcement learning (MARL) is capable of reflecting such worker-to-environment and inter-worker interactions. Depending on the existence of a central controller and the types of interactions, MARL is categorized into centralized/decentralized and cooperative/competitive frameworks, respectively \cite{zhang2019multi}.

%
Centralized MARL frameworks postulate a central controller that learns decision-making polices by collecting all workers' experiences that comprises their observed states, taken actions, and received rewards \cite{khan2018scalable}. 
Exchanging such information may incur huge communication and memory resources while violating data privacy. Decentralized MARL without the central controller does not incur such issues, at the cost of not guaranteeing the equilibrium of the constituted policies of individual workers. Even under cooperative MARL wherein all workers aim to achieve the same goal, it may not guarantee the convergence to equilibrium policies without central coordination \cite{wang2003reinforcement}. Competitive MARL aggravates the problem, wherein every worker's goal competes over a shared common environment and resources as a zero-sum game. Guaranteeing the convergence should thus require additional communication, as we shall discuss with a use case in Sec.~\ref{sec:use_case_MFG}. Nonetheless, note that all the rest of the discussions in this work are centered around distributed learning scenarios that are cooperative and NN based, rather than exploiting MARL in depth.

\section{Key Communication Principles}\label{sec:principles_comm}


Both communication efficiency and accuracy of distributed learning can be improved by leveraging advanced communication principles coping with limited resources and time-varying communication dynamics as discussed in Sec.~\ref{sec:challenges}. Towards improving vanilla distributed learning methods presented in Sec.~\ref{sec:vanilla}, several key communication principles are introduced in this section, and their effectiveness will be elaborated with selected use cases in Sec.~\ref{sec:advanced}.

\subsection{Link Sparsification} \label{sec:principle_sparsification}
Reducing the number of links can significantly decrease the communication bandwidth and energy of distributed learning. Such link sparsification can be implemented in temporal and/or spatial domain. Lazy aggregated gradient descent (LAG) \cite{chen2018lag} is one of its kind pursuing temporal link sparsity by enforcing each worker not to share its model update if the difference, measured by the infinity norm, between the current and previous updates does not exceed a certain threshold. Alternatively, to achieve the spatial link sparsity, one can enforce a sparse network topology by making each worker communicate only with very few neighbors, as exemplified by decentralized gradient descent (GD), dual averaging \cite{duchi2011dual}, and GADMM algorithms \cite{elgabli2020gadmm}.

Link sparsification is not always free, but may come at the cost of higher learning convergence speed and/or lower accuracy. To illustrate, for the spatial link sparficiation, a very sparse network graph (e.g., ring topology with nearest-neighbor based connectivity) yields high communication efficiency per iteration, but may incur more iterations for reaching the convergence and/or a target accuracy level, compared to a denser network graph (e.g., fully connected or star topology with the parameter server). Optimizing the sparsity under the trade-off between per-iteration communication cost and convergence speed is thus crucial.





\subsection{Quantization}\label{sec:principle_quantization}

For each communication round, quantization decreases the number of bits to represent model updates, thereby reducing the communication payload sizes in distributed learning. Due to the reduced arithmetic precision of the model updates, quantization introduces errors, which may hinder the convergence of learning algorithms and/or degrade accuracy. Therefore, a quantizer and its quantizing levels should be carefully designed so as to guarantee the convergence with high accuracy. To this end, one can quantize each element of a gradient vector \cite{alistarh2017qsgd,magnusson2019maintaining,bernstein2018signsgd} or the gradient difference vector between the current and previous model updates \cite{mishchenko2019distributed,chen2018lag}. For the gradient quantization, the methods in \cite{alistarh2017qsgd,magnusson2019maintaining} adjust the qantizing levels under the trade-off between per-iteration communiction cost and the convergence speed. 
SignSGD \cite{bernstein2018signsgd} considers an extreme case wherein gradients are quantized using only $+1$ and $-1$, and shows its convergence by the aid of a majority vote of the workers. There are many other variants of gradient quantized distributed learning algorithms including error compensation \cite{wu2018error}, variance-reduced quantization \cite{zhang2017zipml}, and tenary quantization \cite{wen2017terngrad}.

Quantization can create synergy by integrating with link sparsification elaborated in Sec.~\ref{sec:principle_sparsification}. Lazilly aggregated quantized gradient method (LAQ) is one example that combines the gradient update quantization with temporal sparsification, in a way that the number of links is sparsified based on the temporal gradient update difference, and the gradient update differnece is adaptively adjusted for reducing per-link payload size while ensuring the convergence \cite{chen2018lag}. On the other hand, the method in \cite{elgabli2019qgadmm} merges stochastic quantization with the spatial sparsification of GADMM~\cite{elgabli2020gadmm}, in which the weight update differnece is rounded up and down with probability $p$ and $1-p$, respectively, while $p$ is adaptively adjusted to minimize communication cost while preserving the convergence guarantees of vanilla GADMM~\cite{elgabli2020gadmm}.

The aforementioned methods quantize each element of a model update individually. Alternatively, the model update vector can be quantized altogether by clustering and mapping the updates into the centroids in a multi-dimensional vector space. Leveraging the universal quantization algorithm \cite{zamir1992universal}, the work \cite{shlezinger2020uveqfed} applies universal vector quantization to federated learning, coined UVeQFed, such that the quantization error can be bounded by a term that vanishes as the number of worker grows.



\subsection{Short Packet Aggregation}\label{sec:principle_short_packet}

Whether the length of a communication packet is long or short has a significant impact on communication data rates. To be specific, in a large packet regime, the data rate $R$ can be formalized by the well-known Shannon formula $R=\log(1+\text{SNR})$ per unit bandwidth over the additive white Gaussian noise (AWGN) channel for a given signal-to-noise ratio (SNR). Its derivation relies on assuming an infinite packet length $n\rightarrow \infty$ to ensure the zero packet error probability $\epsilon\rightarrow 0$ \cite{durisi2016toward}, and thus becomes a tight approximation for large packets. Since packet lengths are proportional to communication payload sizes, in the distributed learning context, the Shannon formula is suitable for deep NNs with perodic model parameter exchanging methods such as FL.

By contrast, SL exchanges a single NN layer's instantaneous activation and gradient whose corresponding packet length can be very short. In this short packet regime with finite $n$ and non-negligible $\epsilon$, the data rate $R(n,\epsilon)$  can be described using a formula proposed by Y. Polyanskiy et al. \cite{polyanskiy2010channel}, given as:
\begin{align}
    R(n,\epsilon) = \log(1 + \text{SNR}) -\sqrt{\frac{V}{n}}Q^{-1}(\epsilon) + \mathcal{O}\l(\frac{\log n}{n}\r),
\end{align}
where $Q(\cdot)^{-1}$ is the inverse of the Gaussian Q function, and $V$ is the term capturing channel dispersion, e.g., under the AWGN, $V=(2\text{SNR} + \text{SNR}^2)(\log e)^2/{(1+\text{SNR})^2}$. This formula implies that the short packet length $n$ incurs a penalty on the data rate that is proportional to $1/\sqrt{n}$. To alleviate such a penalty, one can aggregate consecutive packets, increasing $n$ \cite{popovski2018wireless}. Through the lens of SL, this packet aggregation coincides with increasing the batch size of each worker. A larger batch size often yields faster convergence at the cost of compromising accuracy \cite{keskar2016large}. Consequently, there exists a trade-off among data rate, batch size, and accuracy in SL, as we shall discuss in Sec.~\ref{sec:parsl}.




\subsection{Analog Transmission} \label{sec:principle_analog_transmission}

The limited communication bandwidth is one key challenge in distributed learning over wireless channels. The wirelessly connected workers using the same channel may interfere with each other during their over-the-air transmissions. To avoid their interference, under digital transmissions, it is common to avoid such interference by allocate orthogonal channel bandwidths to different workers~\cite{park2018wireless,FL_Nishio,Wang:2019aa,YangQuekPoor:2019aa,Chen:20019aa}. As a result, the workers compete over the limited bandwidth, which is thus not scalable for supporting a large number of workers. Alternatively, motivated by the fact that the parameter server in FL is interested in the aggregated model updates of all workers, i.e., global model $\boldsymbol{\Theta}=\frac{1}{N}\sum_{n=1}^N \boldsymbol{\theta}_n$ with $N$ workers, rather than the individual updates $\boldsymbol{\theta}_n$, several recent works have utilized analog transmissions so as to harness interference without separate channel allocation~\cite{Amiri:SPAWC19,Zhu:19,Sery:19,Zhu:20}.

Under analog transmissions, each transmitted signal from a worker in FL is perturbed by fading, i.e, multiplied by the fading gain $h_n$, and superpositioned over-the-air with all other workers' signals using the same channel. Consequently, $\frac{1}{N}\sum_{n=1}^N h_n \boldsymbol{\theta}_n$ is received by the parameter server. The suporpositioning property of analog transmissions is favorable for averaging the models updates using the entire bandwidth for all workers, rather than competing over the limited bandwidth with each other under digital transmissions. By contrast, the fading perturbation may hinder obtaining the received signal in a desired form at the parameter server, e.g., equal or weighted averaging with the weight that is proportional to the ratio of each worker's data size \cite{Brendan17}. One way to cope with the fading perturbation is the channel inversion method. By inversely perturbing the signal before transmission, i.e., multiplying by $1/h_n$, the fading can be canceled out at reception \cite{amiri2020federated}. This channel inversion however consumes the transmit power inversely proportional to the channel gain, which is not viable for small $h_n$ under the limited edge device energy budget. For this reason, it is common to allow transmissions only when the channel gains exceed a certain threshold \cite{Amiri:SPAWC19,Zhu:19,Sery:19}. As discussed in Sec.~\ref{sec:principle_sparsification}, such temporal sparsification may hinder the convergence of learning algorithms. 

Alternatively, the method proposed in \cite{elgabli2020harnessing} only utilizes the superpositioning property of analog transmissions without channel inversion. This is done by reformulating FL and optimizing it direcly with perturbed model updates as follows. To be specific, recall the original unconstrained problem of FL, aiming to minimize $\frac{1}{N}\sum_{n=1}^N f_n(\boldsymbol{\Theta})$, by locally minimizing $f_n(\boldsymbol{\theta}_n)$ at each worker and globally averaging their model parameters $\boldsymbol{\theta}_n$ at the parameter server, yielding $\boldsymbol{\Theta}$. This boils down to the following constrained average consensus problem:
\begin{align} 
        \min_{\boldsymbol{\Theta},\{\boldsymbol{\theta}_n\}_{n=1}^N}\ \ \ & \sum_{n=1}^N f_n(\boldsymbol{\theta}_n)    \\
    \text{s.t.}\ \ &
        \boldsymbol{\theta}_{n} =\boldsymbol{\Theta}, \ \ \forall n. \label{P1_const}
    \end{align} 
To incorporate the fading perturbed model updates in the problem formulation, by multiplying the fading gain $h_{n}$ at both sizes, \eqref{P1_const} is recast as its equivalent constraint $h_{n}\boldsymbol{\theta}_{n} = h_{n}\boldsymbol{\Theta},  \forall n$. This reformulated problem is solved using ADMM while directly incorporating the perturbed model updates, i.e., $h_{n}\boldsymbol{\theta}_{n}$, without inverting the fading gain $h_{n}$. As a consequence of avoiding channel inversion, the convergence becomes less sensitive to the transmit power constraint. Furthermore, thanks to directly exploiting the perturbed model updates, it is more robust against the adversarial or honest-but-curious parameter server, to be further elaborated in Sec.~\ref{sec:a-fadmm}.

\subsection{Scheduling and Offloading}\label{sec:principle_scheduling}

Heterogeneity is prevalent in distributed learning, in terms of the availability and access to the training data and resources for the communication, computation, and memory. Such heterogeneity results in the learning workers having outdated models compared to other workers, referred to as \emph{stragglers}. Waiting these stragglers may cause significant delays to the overall training operations, whereas ignoring them may hinder guaranteeing the convergence or achieving high accuracy. Scheduling is effective in balancing and resolving this straggler handling problem. To this end, it is of paramount importance to identify the root cause of each straggler and its contribution to the overall learning performance.

The lack of computing resources can be one major cause of stragglers. 
It happens when large-sized models and datasets with multiple tasks are processed by on-device and battery-limited workers. 
In this case, as studied in \cite{amiri2019computation}, an effective solution could be scheduling the resultant stragglers while offloading their computationally demanding tasks (or even training data with a loss of privacy) to neighbors or edge servers, a conceptual design known as mobile edge computing (MEC) \cite{Elbamby:2019,polese2020machine}.
Such task offloading in MEC needs to take into the account of device heterogeneity \cite{zhang2020hetmec}, communication limitations \cite{li2017communication,chanyour2019energy}, and demand-supply capabilities of processing power \cite{wang2019adaptive} in addition to its impact on the tolerable training latency\cite{polese2020machine} and target training/inference accuracy \cite{mohammad2019adaptive} while ensuring devices' privacy \cite{xu2018queryguard}.

Another source of stragglers is poor channel conditions such as  
the channels in deep fades and high interference, as well as communication resource limitation such as limited bandwidth and uplink transmit power. To remedy this type of straggler problem, it is useful to utilize advanced multiple access control techniques such as joint scheduling and resource management, interference mitigation and alignment, proactive scheduling via channel prediction, and multi-hop relaying \cite{yang2019scheduling,wadu2020federated,Han:Intellisys20,abad2020hierarchical}. Reflecting both computing and communication limitations, adjusting the model complexity is also effective in mitigating stragglers \cite{jiang2019model}.



%

%




\section{Key Machine Learning Principles}\label{sec:principles_ml}

Communication efficiency of distributed learning is significantly affected by ML architectures and algorithms. In this section, several machine learning principles are presented for improving vanilla distributed learning methods discussed in Sec.~\ref{sec:vanilla}, and their effectiveness will be validated by representative use cases in Sec.~\ref{sec:advanced}.

\subsection{ {Model Split} }\label{sec:principle_model_split}



Running a large-sized deep NN consumes huge memory that may not fit within edge devices. The energy consumption of this model is proportional to the model sizes \cite{han2015learning}, aggravating the problem under battery-limited edge devices. 
SL resolves such issues by splitting a single NN model into multiple segments stored and operated by different edge nodes. In essence, this problem is traced back to model parallelism focusing on how to partition and offload NN segments, as opposed to data parallelism considering a large-sized global dataset dispersed across different workers running NN models, each of which is separate but has the same architecture \cite{yadan2013multi}.

Traditionally model parallelsm has focused primarily on the NN partitioning based on computing latency \cite{kang2017neurosurgeon}. For instance, a convonlutional NN comprises fully-connected layers and convolutional layers, and the convolutional layers often consume much longer processing delays compared to the fully-connected layers, e.g., in AlexNet \cite{krizhevsky2012imagenet} and ResNet \cite{he2016deep} architectures. Therefore, even if two edge nodes have the same memory size, equally partitioning an NN may not be an optimal way, incurring imbalanced processing overhead. Beyond this, in the context of SL, communication efficiency and data privacy should also be taken into account. Indeed, cutting a NN's bottleneck layer having the smallest dimension (e.g., VAE's bottleneck layer for latent variables \cite{MAL-056}) can maximally reduce the SL communication payload sizes. Furthermore, in a classification task, not only unlabled data samples but also their ground-truth labels can be privacy-sensitive (e.g., unlabled X-ray images and their ground-truth diagnosis results) \cite{jeong2018communication,jeong2019multi,dsn19}. 
In this case, the input and output layers are linked to the raw samples and ground-truth labels (for training loss calculation), respectively, and a NN should thus be partitioned such that the input and output layers are colocated at the data owner. More discussions on model split are deferred to Sec.~\ref{sec:medicalsl}.


\subsection{{Knowledge Distillation}} \label{sec:principle_KD}
%
Knowledge distillation (KD) aims to imbue an empty student model with a teacher's knowledge \cite{HintonKD:14}. In a classification task, as opposed to the standard model training that attempts to match the student model's one-hot prediction (e.g., [cat, dog] = [0,1]) of each unlabled sample with its ground-truth label, KD tries to match the model's output layer activation, so-called logit (e.g., [cat, dog] = [0.3, 0.7]), with the teacher's logit for the same sample. This logit contains more information than its one-hot prediction, thereby training the student model faster than the standard training with much less samples \cite{Phuong19}.

The teacher's knowledge of KD can be constructed in different ways. Originally, the knowledge is a pretrained teacher model's logit, which is transferred to a smaller student model for model compression \cite{HintonKD:14}. 
The knowledge can also be an ensemble of other student models' logits \cite{anil2018large}, in that the ensemble of predictions is often more accurate than individual predictions. Leveraging this to enable distributed learning, the knowledge in FD is constructed the ensemble of different workers' prediction, each of which is locally averaged per label in a classification task \cite{jeong2018communication,ahn2019wireless} or across neighboring states in reinforcement learning \cite{cha2019federated}. The local averaging step avoids the same sample observations of the student and teacher models (i.e., ensemble of all student models), thereby reducing significant communication overhead while preserving local data sample privacy. Lastly, for given averaged logits as the teacher's knowledge, running KD with an empty student model at the parameter server realizes a fast one-shot FL or the information type conversion from logits to the parameters of the trained student model, which will be discussed with a use case in Sec.~\ref{fig:usecase_Mix2FLD}.

%

\subsection{Mixup Augmentation}\label{sec:principle_data_aug}




Mixup is a data augmentation technique generating a synthetic sample by superpositioning two different samples \cite{Zhao2018}. As an example, in a binary classification task, a sample~$s_0$ in the label $0$ is linearly combined with another sample~$s_1$ in the label $1$, thereby yielding a synthetic sample $\hat{s}_{01}$ given~as:
\begin{align}
    \hat{s}_{01}= \lambda s_0 + (1-\lambda) s_1.
\end{align}
The term $\lambda$ is the mixing ratio that is randomly sampled from a bathtub-shaped beta distribution such that $\hat{s}_{01}$ resembles a sample in the  label either $0$ or $1$ with a slight difference. Manifold Mixup applies the same technique to superposition two different hidden representations, which often performs similar or even higher accuracy than vanilla Mixup that combines raw samples \cite{verma2019manifold}.

Both vanilla and Manifold Mixup are commonly used in standalone training, particularly for adversarial learning that intentionally feeds distorted samples to obtain more generalized models \cite{Zhao2018,verma2019manifold}. In distributed learning, these techniques can also be utilized for sharing proxy samples without revealing raw data samples \cite{Park:2018ab}. 
For example, to rectify non-IID data distributions, each worker can exchange mixed-up samples or manifold mixed-up representations to complement missing samples in some labels \cite{jeong2019multi}. 
By uploading the mixed-up samples or representations to a parameter server, the workers' training computation can be offloaded to the server enabling one-shot FL \cite{shin2020xor}. 
The number of these generated proxy samples or representations can further be oversampled by mixing them across different workers \cite{cha2019federated} and/or re-mixing the mixed-up samples or representations \cite{Oh20:CL}. 
More use cases and effectiveness of Mixup and manifold Mixup will be discussed in Sec.~\ref{sec:mix2fl}, \ref{sec:xorfl}, and \ref{sec:heteromodalsl}.

\subsection{Gaussian Process Regression}\label{sec:principle_gpr}

%
Dynamics in the environment, agents’ hardware, and random choices of training batches and learning parameters cause computing and communication resources and training model parameters to change over the training duration. 
These dynamics of resource and model states can be viewed as stochastic processes.
Considering a Gaussian process prior probability distribution on above stochastic processes provides means of analyzing them using Bayesian inference methods \cite{rasmussen2003gaussian}.
Gaussian process regression (GPR) is the process of determining a set of kernel hyperparameters defining the covariance matrix between the all possible observations over time (and space) assuming zero-mean distribution therein.
Using GPR, the posterior mean and variance at unseen observations can be analytically estimated.
By modeling the dynamics of the resource (computation and/or communication) availability as a time series, GPR can be adopted to predict future resources (mean) with the uncertainty bounds (variance) \cite{girard2003gaussian}.
This allows to identify agents who are likely to be stragglers in advance, in which agents and resources can be proactively scheduled.
As a result, the overall training latency can be decreased with minimum loss of training performance and the overall resource utilization can be improved.
Similarly, model parameter dynamics can be analyzed with GPR. 
Under the communication bottleneck in collaborative learning, agents can utilize estimated model parameters of others to continue local training while using the limited resources only when the uncertainties of model estimations are unacceptable.  

\subsection{ {Mean-Field Game (MFG) Learning}}\label{sec:principle_MFG}

Decentralized decision-making of competitive and mutually interactive workers is a challenging task as discussed in Sec.~\ref{sec:method_MARL}. Due to these interactions, it is common to determine a single worker's action by fixing all other worker states, and then iterate it for the next worker until all workers' actions converge to the Nash equilibrium, a stable state at which no worker gains more reward by changing its action \cite{kreps1989nash}. 
The complexity of this problem is thus increasing exponentially with the number of workers, which is unfit for dealing with massive interactive workers. Mean-field game (MFG) is a useful framework to greatly reduce the complexity \cite{MobilMFGSG:GC16,kim:2017:MFCA,KimSPAWC:18,shiri2019massive,shiri2020communication}. At its core, MFG approximates the problem of massive interactive workers as the problem of each single worker interacting with a virtual worker whose state distribution is given by the distribution of the entire population. Then each worker's decision-making boils down to solving two partial differential equations (PDEs), the Hamilton-Jacobi-Bellman (HJB) equation and the Fokker-Plank-Kolmogorov (FPK) equation \cite{lasry2007mean}. 
By solving FPK, one can obtain the population state distribution, called mean-field (MF) distribution. For the given MF distribution, solving HJB results in the optimal action of each worker. 

One common limitation of MFG-theoretic approaches is the curse of dimensionality, which is detoured by the MFG learning framework. To be specific, a PDE is often solved numerically by discretizing the domain so that the derivatives therein can be approximated using finite differences. To guarantee the convergence of such a finite difference method, the discretizing step size should decrease with the domain dimension. As an example, for a given $n$-dimensional domain vector $\{x_1,x_2,\cdots,x_n\}$, the discretization step size $\Delta t$ should satisfy $\Delta t \leq (\sum_{i=1}^n 1/x_i)^{-1}$ according to the Courant-Friedrichs-Lewy condition \cite{courant1967partial}. 
Consequently, the dimensionality increase in states and actions incurs huge extra computing overhead for solving FPK and HJB equations, respectively. MFG learning resolves this issue by recasting the problem of solving HJB and FPK equations, i.e., $H=0$ and $F=0$, respectively, as the regression tasks of minimizing $|H|$ and $|F|$, respectively. To solve these two regression tasks, a pair of HJB NN and FPK NN are introduced in that NNs are good at tackling regression problems via simple first-order algorithms such as the gradient descent method. The effectiveness of MFG learning will be corroborated with a massive drnoe control use case in Sec.~\ref{sec:use_case_MFG}.




\section{Use Cases: Communication-Efficient and Distributed Learning Frameworks}\label{sec:advanced}


By applying the ML and communication principles introduced in Sec.~\ref{sec:principles_comm} and \ref{sec:principles_ml} to vanilla distributed ML methods in Sec.~\ref{sec:vanilla}, in this section we present communication-efficient and distributed learning frameworks with selected use cases. The mapping between specific principles and use cases is illustrated in Fig.~\ref{fig:intro_overview}.

\subsection{Quantized-GADMM (Q-GADMM)}
\label{sec:qgadmm}


Utilizing GADMM that exploits sparse connectivity (Sec. \ref{sec:principle_sparsification}), Q-GADMM allows each worker to share a quantized version of its model with neighbors \cite{elgabli2019qgadmm}. 
Using stochastic quantization, one of the key communication principle in Sec. \ref{sec:principle_quantization}, with adjustable quantization range, Q-GADMM can significantly reduce the communication energy compared to original GADMM at a zero cost in terms of the convergence speed and accuracy.

The stochastic quantization places the $i$-th dimensional element {$[\boldsymbol{\hat\theta}_n^{k-1}]_i$} of the previously quantized model vector at the center of the quantization range {$2 R_n^k$} that is equally divided into $2^{b}-1$~quantization levels.
This yields a quantization step size $\Delta_n^k=2 R_n^k/(2^{b}-1)$ of resolution $b$.
%
Each worker quantizes the difference between the current and the previously quantized models by choosing a rounding probability yielding a zero quantization error on average.  
%





%

\begin{figure}[!t]
    \centering
    \includegraphics[width=.9\columnwidth]{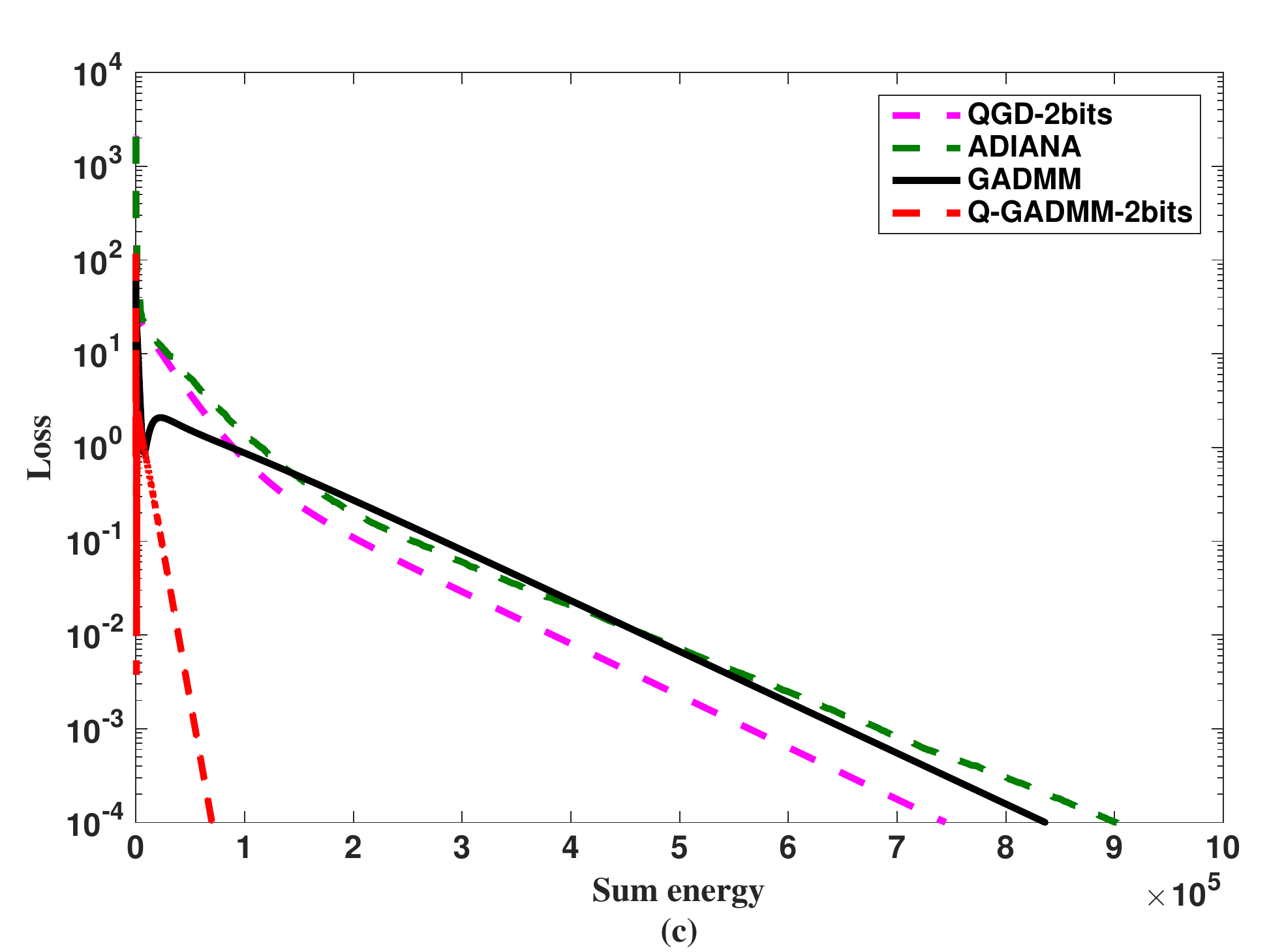}
    \caption{Q-GADMM: relation between energy consumption and relative linear regression loss $(|F-F^*|)$.}
    \label{fig:usecase_Q-GADMM}
\end{figure}

Each worker then transmits $R_n^k$ and the index of the quantization level $q_n(\boldsymbol{\theta}_n^k)$ to its neighboring workers. 
At the receiver, $\boldsymbol{\hat\theta}_n^k$ can be reconstructed by
$\hat{\boldsymbol\theta}_n^k = \boldsymbol{\hat\theta}_n^{k-1}+ \Delta_n^k q_n(\boldsymbol\theta_n^k)-R_n^k\mathbf{1}$.
Consequently, when the full arithmetic precision uses $32$\,bits to represent $R_n^k$, the payload size of Q-GADMM is $(b d + 32)$\,bits where $d$ is the model size.
Compared to GADMM whose payload size is $32 d$\,bits, Q-GADMM can achieve a huge reduction in communication overhead, particularly for large $d$.

\begin{figure}
    \centering
    \subfigure[Communication iterations.]{\includegraphics[width=.9\columnwidth]{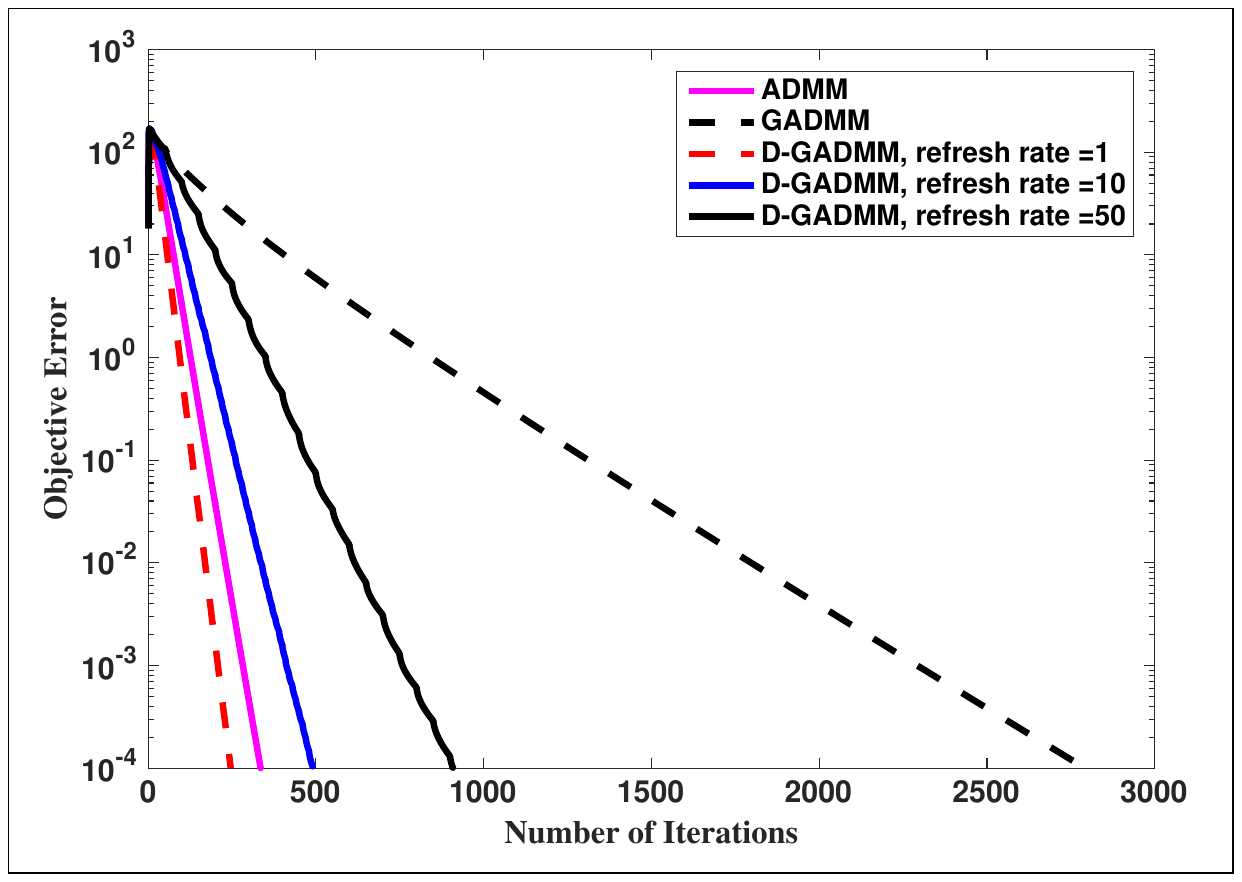}}
    \subfigure[Communication cost.]{\includegraphics[width=.9\columnwidth]{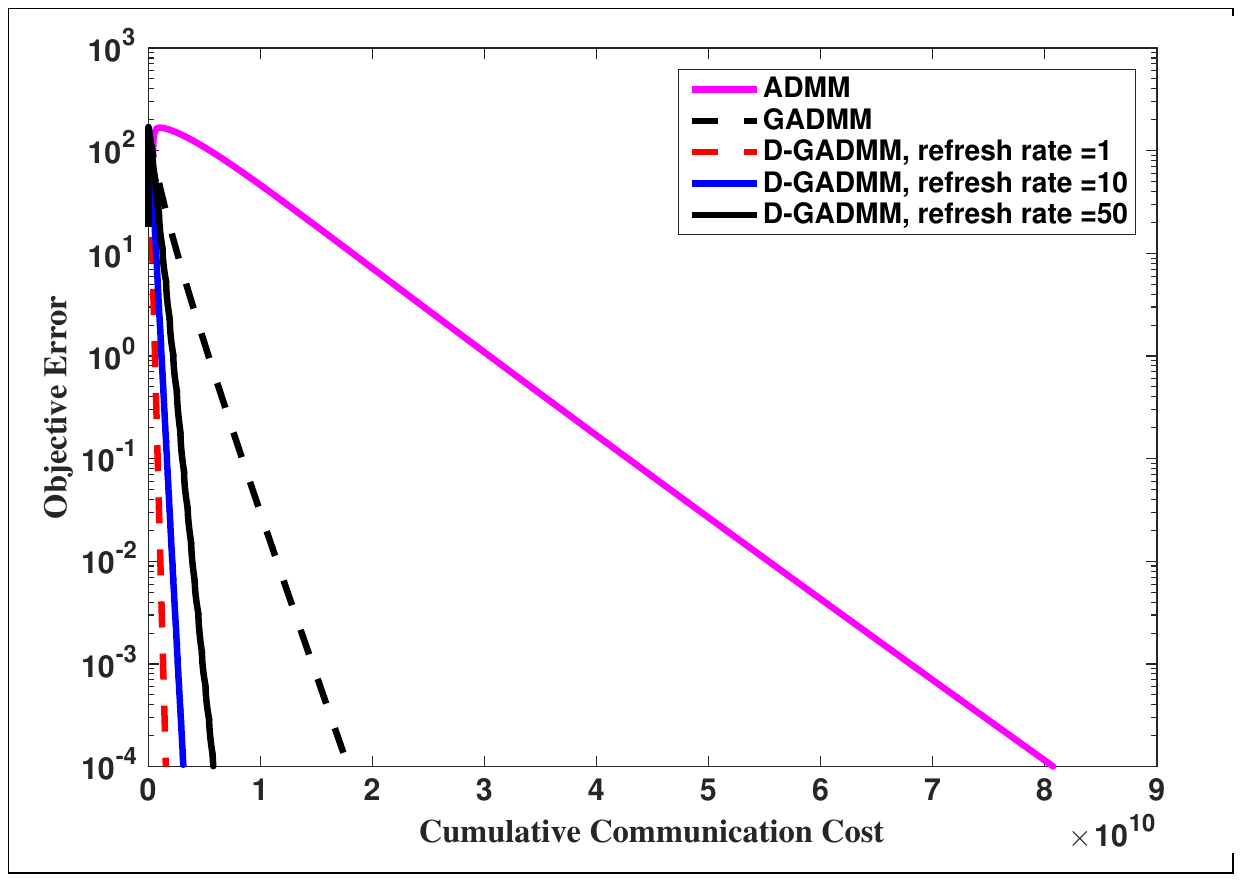}}
    \caption{D-GADMM: comparison of objective errors as functions of (a) communication iterations and (b) communication cost.}
    \label{fig:usecase_D-GADMM}
\end{figure}

Fig. \ref{fig:usecase_Q-GADMM} compares Q-GADMM with GADMM, and two PS-based schemes (QGD, and ADIANA \cite{li2020acceleration}) in terms of the loss versus the total sum energy for a system of $50$ workers.
Here, linear regression of California housing dataset with $d=6$ input features is tested.
%
In the full precision GADMM, each worker will transmit $32d$\,bits to represent all elements in the model vector. 
In contrast, each worker of Q-GADMM only uses $(32+2d)$\,bits, with $2$\,bits to represent each element in the model vector. 
Following the Shannon's capacity theorem, more bits consumes more transmission energy for the same bandwidth, transmission duration, and noise spectral density. 
%
Fig. \ref{fig:usecase_Q-GADMM} exhibits significant reduction in the total energy consumption, a key challenge discussed under Sec. \ref{sec:challenges}, for Q-GADMM compared to all baselines, owing to 
i) the decentralization where workers communicate with only nearby neighbors (Sec. \ref{sec:principle_sparsification}), 
ii) the fast convergence inherited from GADMM (Sec. \ref{sec:method_gadmm}), and 
iii) the reduction of transmitted bits at every iteration while ensuring convergence via stochastic quantization (Sec. \ref{sec:principle_quantization}),.

\subsection{Dynamic GADMM (D-GADMM)}



In practise, due to device mobility, the network topology is time variant, in which neighboring nodes continuously change over time as highlighted in Sec. \ref{sec:challenges}.
Hence, to enable distributed learning over dynamic network of workers, Dynamic GADMM (D-GADMM), which inherits the theoretical convergence guarantees of GADMM is proposed in \cite{elgabli2020gadmm}.
%
%
While adapting to network dynamics, D-GADMM improves the convergence speed of GADMM, i.e.,
random changes in sparse and logical neighbors (Sec. \ref{sec:principle_sparsification}) of a static physical topology can significantly accelerate the convergence of GADMM.
Although the sparsity of network graphs yields slow convergence speeds~\cite{nedic2018network}, the reductions of convergence speed in D-GADMM compared to the standard PS-based ADMM can be compensated by continuously altering neighbors with D-GADMM.
%
%
In addition, with dynamic topology changes, D-GADMM exhibits significant communication cost reductions compared to GADMM \cite{elgabli2020gadmm}. 
%

Fig. \ref{fig:usecase_D-GADMM} compares D-GADMM with both GADMM and standard ADMM. 
From Fig.\ref{fig:usecase_D-GADMM}, it can be seen that utilizing D-GADMM significantly increases the convergence speed of GADMM and hence, reduces the total communication cost even when the topology is fixed. 
Therefore, D-GADMM can compensate for the decrease in the convergence speed of GADMM compared to PS-based ADMM due to topology decentralization and maintains a low communication cost per iteration gained by GADMM.

\begin{figure}[!t]
    \centering
    \includegraphics[width=.9\columnwidth]{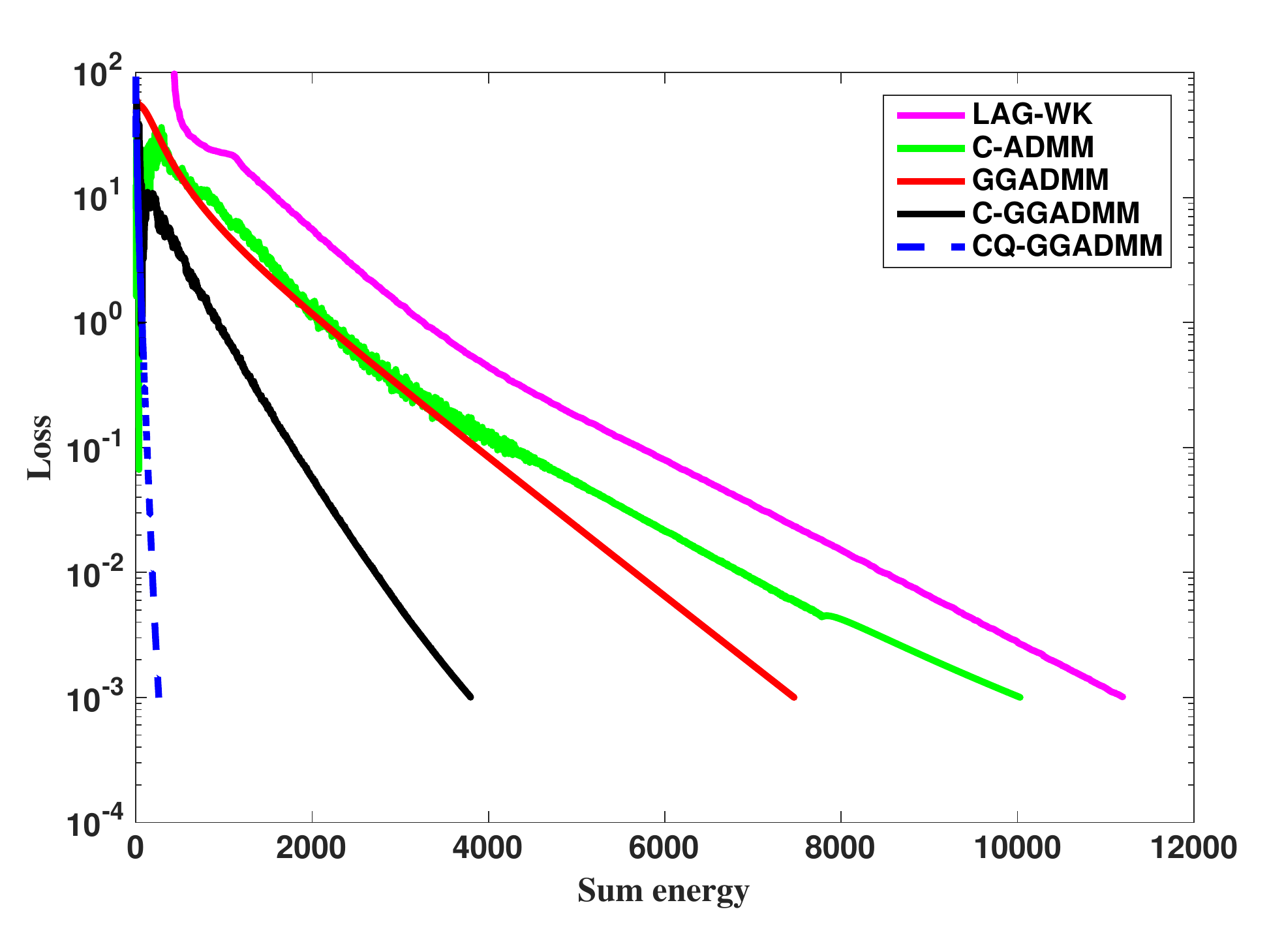}
    \caption{C-GADMM: loss as a function of total energy consumption.}
    \label{fig:usecase_C-GADMM}
\end{figure}

\subsection{Censored Generalized GADMM (C-GGADMM)}




In GADMM, every worker has to share its own model with only up to two neighboring workers at every iteration. To reduce communication overhead while addressing more genereral network topologies, we propose censored generalized GADMM (C-GGADMM). In C-GGADMM, by exploiting temporal sparsity, each worker shares its model only if the difference between the current and the previous models exceeds a certain threshold \cite{chen2018lag}. Furthermore, each worker in C-GGADMM can communicate with an arbitrary number of neighbors in a different group (i.e., under any bipartite graph), which is helpful addressing time-varying network topologies (Sec. \ref{sec:challenges}). Theoretically, C-GGADMM inherits the same performance and convergence guarantees of Vanilla GGADMM, under a non-increasing and non-negative censoring threshold sequence; particularly if the threshold at iteration $k$ follows $\tau_k=\omega \zeta^k$ where $\omega \geq 1$ and $0<\zeta^k<1$. Furthermore, by integrating Q-GADMM and C-GGADMM, we propose C-QGGADMM that performs the censoring based link sparsification with payload quantization (Sec. \ref{sec:principle_quantization}). Consequently, C-QGGADMM decreaes both the cost per channel use and the number of channels, thereby significantly reducing the communication energy and competition on the limited bandwidth.

The benefits of censoring and quantization in terms of reduced energy consumption are elaborated in Fig.\ref{fig:usecase_C-GADMM} using the linear regression problem described in Sec. \ref{sec:qgadmm}. It can be noted that introducing censoring on top of GGADMM can provide about two-fold reduction in the total communication cost. Moreover, implementing both censoring and quantization can further lower the total communication~cost.

\subsection{Analog Federated ADMM (A-FADMM)} \label{sec:a-fadmm}



\begin{figure}[!t]
    \centering
    \subfigure[Linear regression.]{\includegraphics[width=\columnwidth]{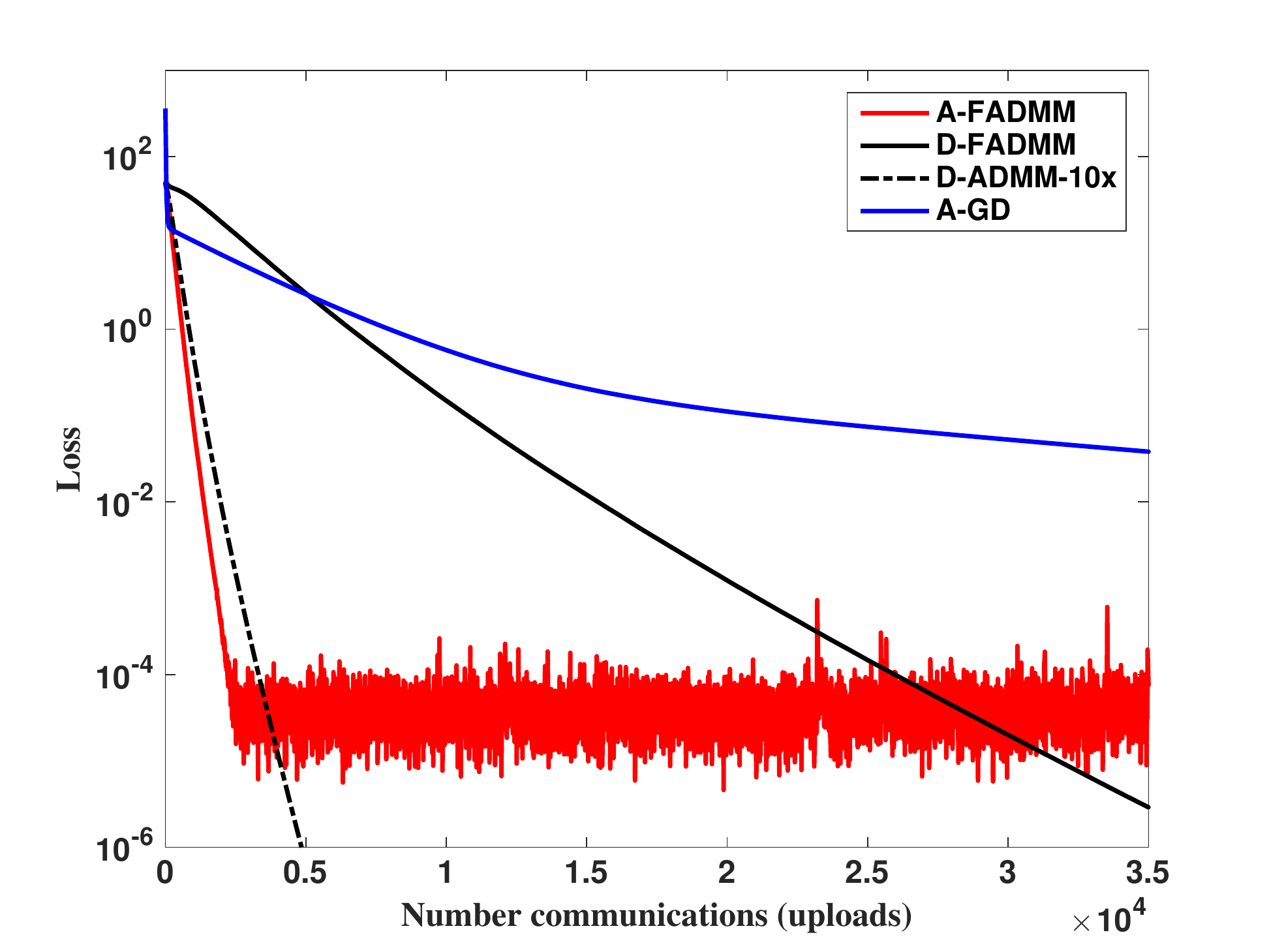}\label{fig:usecase_A-FADMM_linearReg}
    }
    \subfigure[Image classification with a deep NN.]{\includegraphics[width=.9\columnwidth]{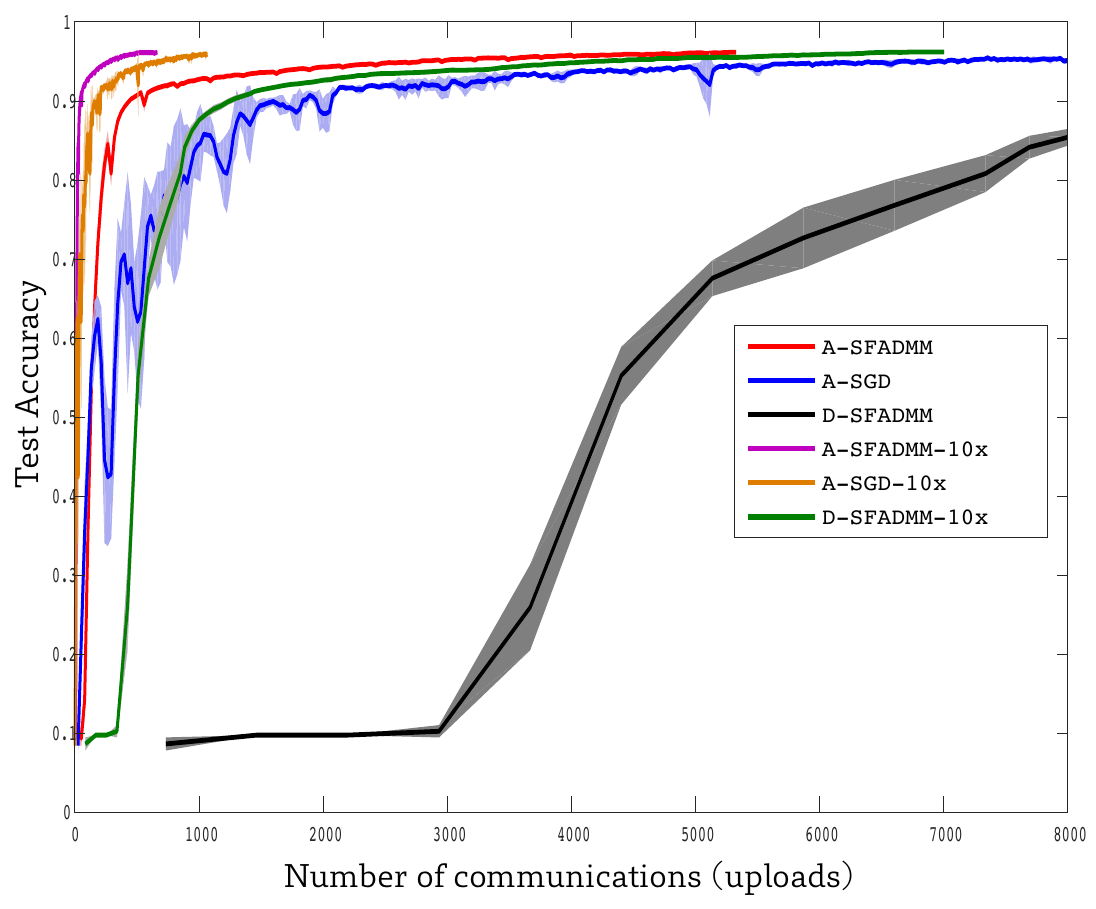}\label{fig:usecase_A-FADMM_dnn}
    }
    \caption{A-FADMM: performance comparison in (a) a linear regression task and (b) the MNIST classification task with a deep NN.}
\end{figure}

In A-FADMM \cite{elgabli2020harnessing}, each worker transmits an analog signal (Sec. \ref{sec:principle_analog_transmission}) that is a function of the $i$-th element in the model over a \emph{shared channel among all workers}. All transmitted signals are superpositioned over the air while hiding each private local model in the crowd preserving the privacy (Sec. \ref{sec:challenges}).
Consequently, the PS receives aggregated signals of all individuals \emph{perturbed} by their complex fading channels. Hence, A-FADMM aggregates multiple workers' updates at the PS without competition on the available bandwidth via analog transmissions. It was proven in \cite{elgabli2020harnessing} that A-FADMM converges to the optimal solution for convex functions and preserves privacy. Moreover, A-FADMM copes with the nuisances incurred by analog transmissions, in terms of time-varying channel fading, noise, and transmit power limitation.

Fig.~\ref{fig:usecase_A-FADMM_linearReg} compares analog and digital implementations of ADMM on a linear regression task. We plot the loss vs the number of uploads (communication rounds). As observed in Fig.~\ref{fig:usecase_A-FADMM_linearReg}, A-FADMM requires the lowest communication rounds to achieve a target loss $10^{-4}$. 
Even with $10\times$ more subcarriers, D-FADMM fails to reach the same speed due to the orthogonal subcarrier allocation to each worker under limited bandwidth. However, if one aims to achieve very low loss below $10^{-4}$, A-FADMM suffers from noisy reception, and D-FADMM may thus be a better choice, as long as very large bandwidth and/or long uploading time are available.

Fig.~\ref{fig:usecase_A-FADMM_dnn} validates the applicability of the stochastic version of A-FADMM (A-SFADMM) on the stochastic and non-convex problem of image processing using DNN. Note that the model size for the tested DNN architecture is several order of magnitudes higher than the model size of the linear regression problem discussed above (For the simulation details see \cite{elgabli2020harnessing}). As observed from Fig.~\ref{fig:usecase_A-FADMM_dnn}, A-SFADMM significantly outperforms the digital implementation (D-SFADMM) in terms of the convergence speed while achieving the maximum accuracy. In fact, A-SFADMM outperforms 10x-D-SFADMM which has $10\times$ larger badnwdith (i.e., $10\times$ more subcarriers).

\subsection{Quantum Scheduler Aided FL}

\begin{figure}[!t]
	\centering
	\subfigure[QAOA quantum circuit.]{
		\includegraphics[width=\columnwidth]{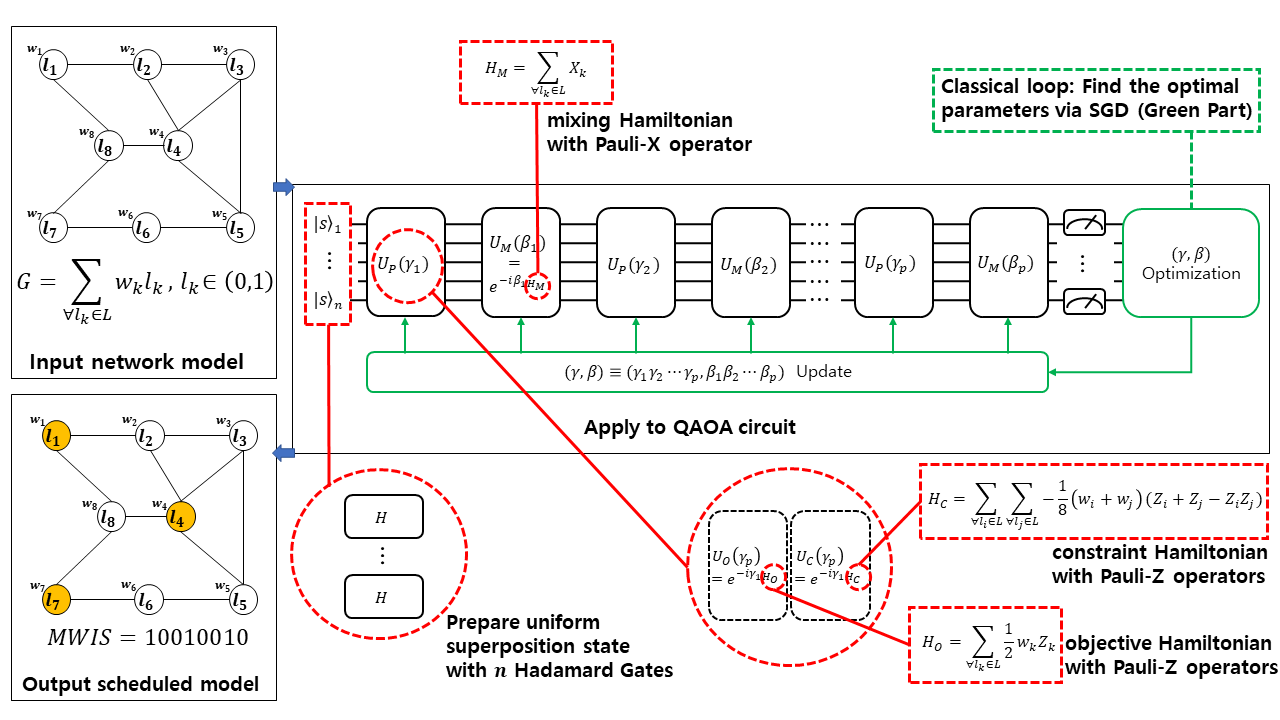}
		\label{fig:quantumfl_qaoa}
	}\vspace{.3cm}
	\subfigure[Experiment trials and their related performance CDFs.]{
		\includegraphics[width=\columnwidth]{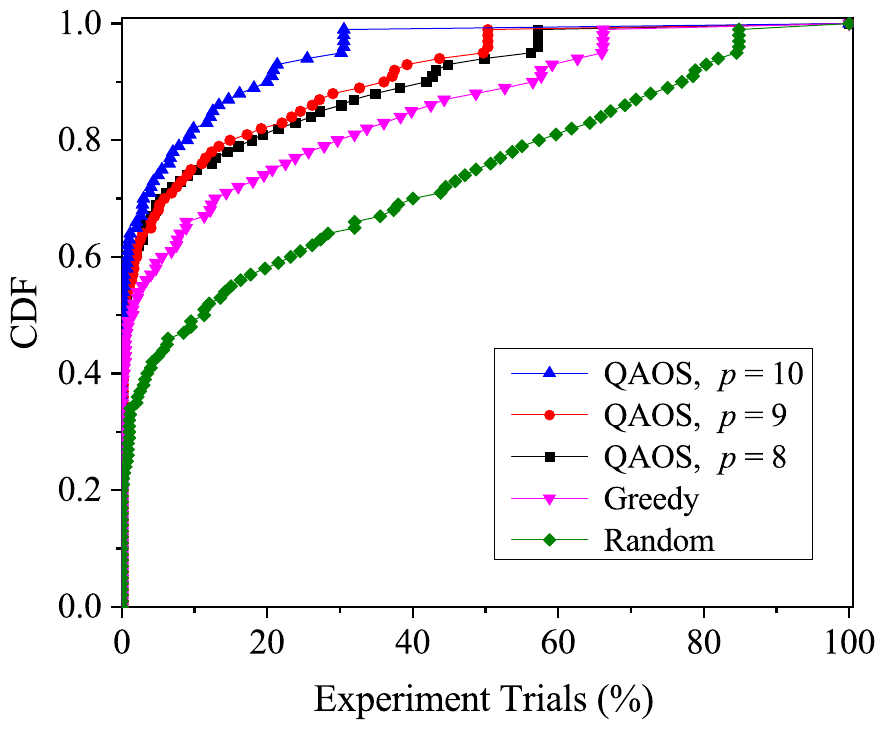}
		\label{fig:usecase_QFL}
	} 
	\caption{QAOA-based approach for an MWIS scheduling problem: (a) a schematic illustration of the QAOA quantum circuit, and (b) the CDF of the ratio between the QAOA-scheduled traffic of $10$ workers and the traffic scheduled via an exhaustive search.}
	\label{fig:QAOA}
\end{figure}

In modern quantum computing research, the design and implementation of quantum approximate optimization algorithms (QAOA) is of great interest \cite{farhi2014quantum,zhou2018quantum}. 
With QAOA-based methods, many approximation algorithms for NP-hard problems are under development. 
Among the NP-hard problems, the QAOA-based approximation solution approach to max-weight independent set (MWIS) problem is actively under discussion where the MWIS formulation is widely used for network scheduling modeling, e.g., device-to-device wireless networks~\cite{ton16}. 
As studied in~\cite{Ma2020SchedulingPA}, scheduling problems are considered and formulated with MWIS in FL over wireless channels where the objective for the scheduling is sum-rate-maximization. 

This QAOA-based approach finds proper parameters for quantum approximation from classical optimization approaches.
The use of QAOA is beneficial in terms of computation time and complexity comparing to the other MWIS solution approaches such as message-passing. Based on the parameters, the approximation solution to the given problem can be obtained from the optimum of the expectation value of Hamiltonian. 
As presented in Fig.~\ref{fig:usecase_QFL}, the QAOA-based MWIS schedulers outperform greedy and random scheduling baselines, wherein the performance is measured by the cumulative distribution function (CDF) of the proportion between the QAoA-scheduled workers' weights and the optimal weights after an exhaustive search. The same tendency holds for various $p$ values, where $p$ means the number of alternation of quantum approximation computation, having an impact on the convergence speed and accuracy. Fig.~\ref{fig:quantumfl_qaoa} illustrates the $p$-level quantum circuit under study, describing the QAOA quantum gate computing procedures for solving the MWIS problem. The QAOA-based quantum scheduler can be designed and implemented using Cirq and TensorFlow-Quantum~\cite{tfq}, where Cirq is a Python framework for creating, editing, and invoking noisy intermediate scale quantum (NISQ) circuits, while TensorFlow-Quantum integrates quantum computing algorithm with the logic designed in Cirq.

\begin{figure}
    \centering
    \includegraphics[width=.9\columnwidth]{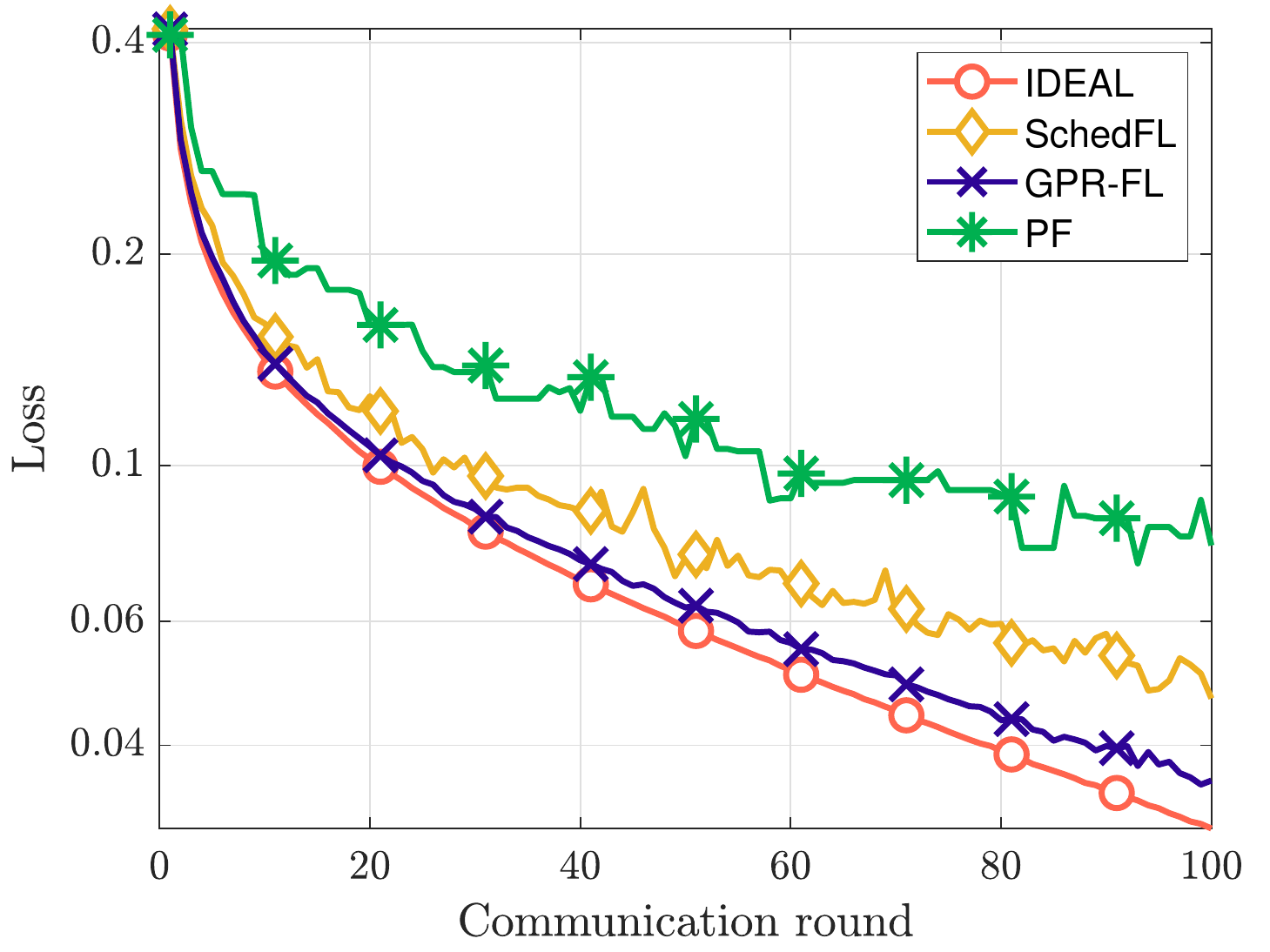}
    \caption{Comparison of FL over MNIST dataset with GPR-based channel prediction and joint agent and wireless resource scheduling.}
    \label{fig:usecase_GprFL}
\end{figure}

\subsection{GPR Aided FL}



%
Communication plays a key role in the local model aggregation and global model sharing steps of the FL (Sec. \ref{sec:method_FL}) over wireless networks.
The poor channel conditions in both uplink and downlink introduces stragglers from the communication point of the view (Sec. \ref{sec:challenges}), in which, channel measurement or accurate estimation is essential under the limited communication resources \cite{kairouz2019advances}.
Although measuring channels aids to utilize agent and resource scheduling, the channel sampling and pilot transmissions therein require high reliable (possibly dedicated) resources as well as introduce significant latency to the training process.
To overcome the cons of channel measurement, GPR-based channel estimation can be adopted in FL (GPR-FL) \cite{wadu2020federated}.
By modeling the dynamic channel states as stochastic processes with a Gaussian prior, time series prediction in GPR can be used to estimate the channels and their uncertainty (Sec. \ref{sec:principle_gpr}).
Using the uncertainty of channels from GPR as a regularizer within FL loss function, joint channel sampling and allocation for straggler-free scheduling (Sec. \ref{sec:principle_scheduling}) can be carried out simultaneously to reduce the sampling latency \cite{karaca2012smart}.
To illustrate the benefits of GPR-FL, we compare the training loss dynamics (relative to the loss of centralized training) of GPR-FL under limited wireless resources with three other methods as illustrated in Fig. \ref{fig:usecase_GprFL}:
i) \emph{SchedFL}: joint agent and resource scheduling towards minimizing training loss similar to GPR-FL is used with channel measurements, 
ii) \emph{PF}: proportional fair scheduling in terms of contribution to model aggregation without channel measurements, 
and iii) \emph{IDEAL}: FL without communication constraints.
Note that a single resource block is dedicated for the channel measurement in SchedFL.
GPR-FL reaps the benefit of the additional resource by utilizing it in agent scheduling over SchedFL, yielding a lower loss as close to IDEAL.
In contrast, PF performs poorly even with the additional resource, due to the absence of the training loss minimization objective within its scheduling policy. 
It is worth noting that due to underlying complexity in GPR and lack of channel sampling, GPR-FL may tend to lose its performance under the availability of excessive amount of total resources compared to the number of agents.
A viable solution is to limit agents’ access to subsets of resources rather the entire resource pool.
%




\subsection{Federated MFG Learning for Massive UAV Control} \label{sec:use_case_MFG}

By integrating MFG learning with FL, in this use case we study controlling a massive number of unmanned aerial vehicles (UAVs) in a communication-efficient and decentralized way. Following the MFG learning framework as elaborated in Sec.~\ref{sec:principle_MFG}, each UAV is equipped with a pair of HJB and FPK NNs. The HJB NN outputs (i) the UAV's optimal action (i.e., acceleration) and (ii) the resultant cost functional value by feeding (iii) the UAV's observed state and (iv) the state distribution of the entire UAV population (i.e., MF distribution). The FPK NN outputs (iv) the MF distribution by feeding (iii) the UAV's state and (ii) the cost functional value obtained from the HJB. While (iii) is fixed, (ii) and (iv) are recursively updated until convergence, at which the optimal action (i) is finally determined \cite{shiri2019massive,shiri2020communication}. 
According to the MFG theory \cite{Bensoussan:2013:MFG}, the aforementioned optimal control can achieve the epsilon-Nash equilibrium as long as the initial states of all UAVs are exchanged without any further inter-UAV communication. This is true when the outputs of the HJB and FPK NNs accurately approximate the solutions of the HJB and FPK equations; in other words, HJB and FPK NNs are ideally trained, which is not feasible due to the lack of training samples (i.e., observed states). 

\begin{figure}
    \centering
    \includegraphics[width=.95\columnwidth]{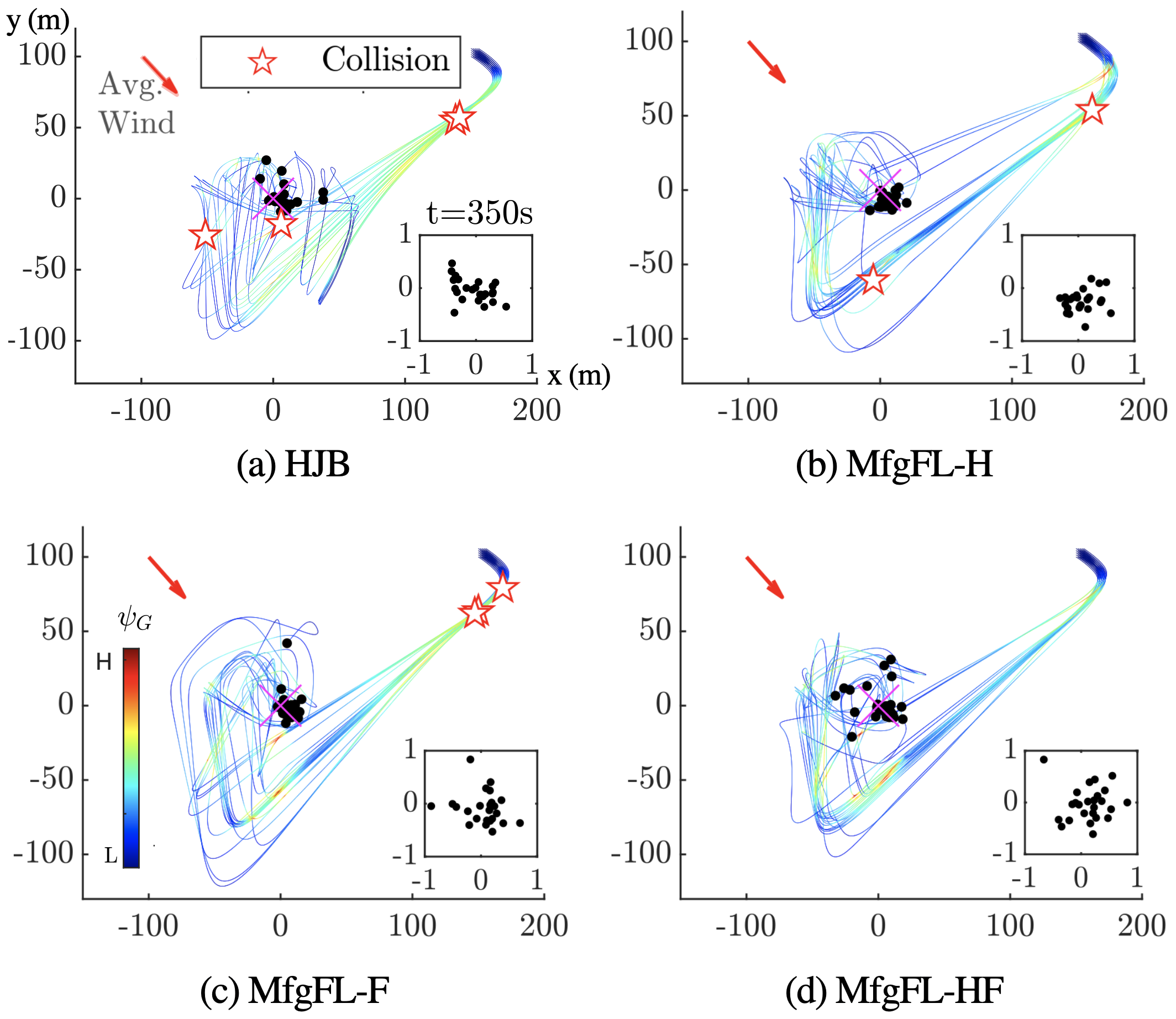}
    \caption{Trajectories of 25 UAVs dispatched from a common source (top right) to a destination (bottom left), when each UAV runs MFG learning while exchanging: (b) only HJB NNs, (b) only FPK NNs, and (c) both HJB and FPK NNs with neighbors, compared to a baseline in which (a) each UAV runs only an HJB NN while exchanging raw states.}
    \label{fig:usecase_mfgfl-trajectory}
\end{figure}

\begin{figure}
    \centering
    \includegraphics[width=\columnwidth]{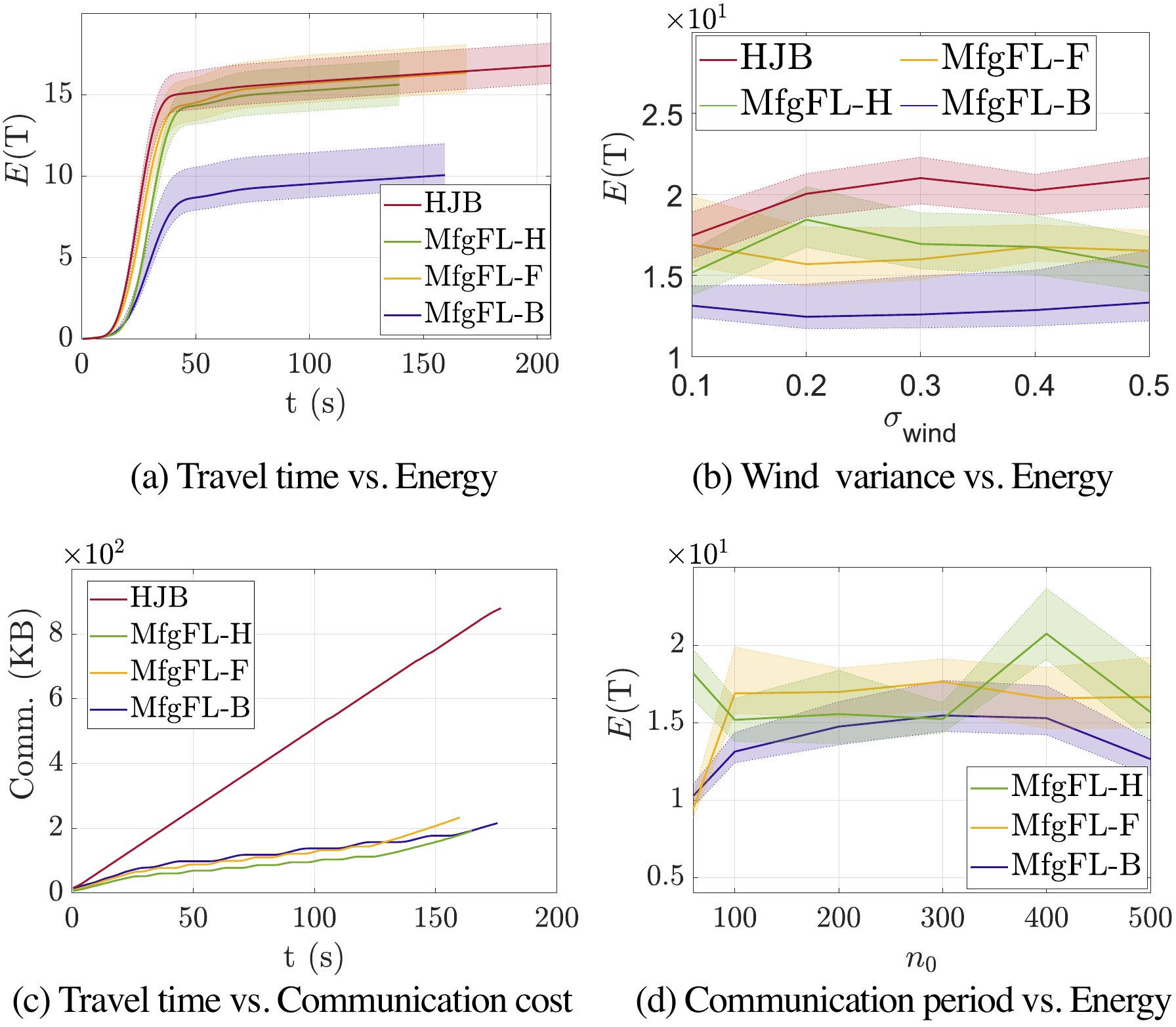}
    \caption{Performance comparison of MfgFL in terms of: (a) energy and travel time, (b) energy for various wind velocity variance values, (c) accumulated communication cost during travel, and (d) energy for various communication periods.}
    \label{fig:usecase_mfgfl-sim}
\end{figure}

To accelerate the training of HJB and FPK NNs, following FL, each UAV periodically broadcasts its NN weights with its neighbors, and updates its model by averaging the received weights within a predefined latency deadline. 
As each UAV has HJB and FPK NNs, there are three possible configurations, exchanging only HJB NN (MfgFL-H), only FPK NN (MfgFL-F), or both HJB and FPK NNs (MfgFL-B) at the cost of the increased communication payload sizes. With 25 UAVs dispatched from a common source to a destination, Fig.~\ref{fig:usecase_mfgfl-trajectory} shows that MfgFL-B achieves the best trajectory without any collision, while all MfgFL based methods yield better results than a baseline operates by only running the HJB NN while exchanging raw states of neighboring UAVs. Here, the curve color indicates the value of $\phi_G$ in the cost function, a swarming term that decreases with the relative velocities, and increases with the relative distances of all UAVs. Again, MfgFL-B yields the lowest $\phi_G$ even at the early stage, supporting the collision-free results. Furthermore, MfgFL-B consumes the minimum motion energy until reaching the destination as shown in Fig.~\ref{fig:usecase_mfgfl-sim}(a), and is more robustness against external disturbances reflected by the variance of random wind velocity as observed in Fig.~\ref{fig:usecase_mfgfl-sim}(b). In addition, Fig.~\ref{fig:usecase_mfgfl-sim}(c) illustrates that MfgFL-B exchanges the least amount of the packets even though its per-communication payload size is 2x greater than MfgFL-H or MfgFL-B. Lastly, for different communication periods, MfgFL-B results in the least energy consumption as seen by~Fig.~\ref{fig:usecase_mfgfl-sim}(d).

\begin{figure}
    \centering
    \includegraphics[width=.9\columnwidth]{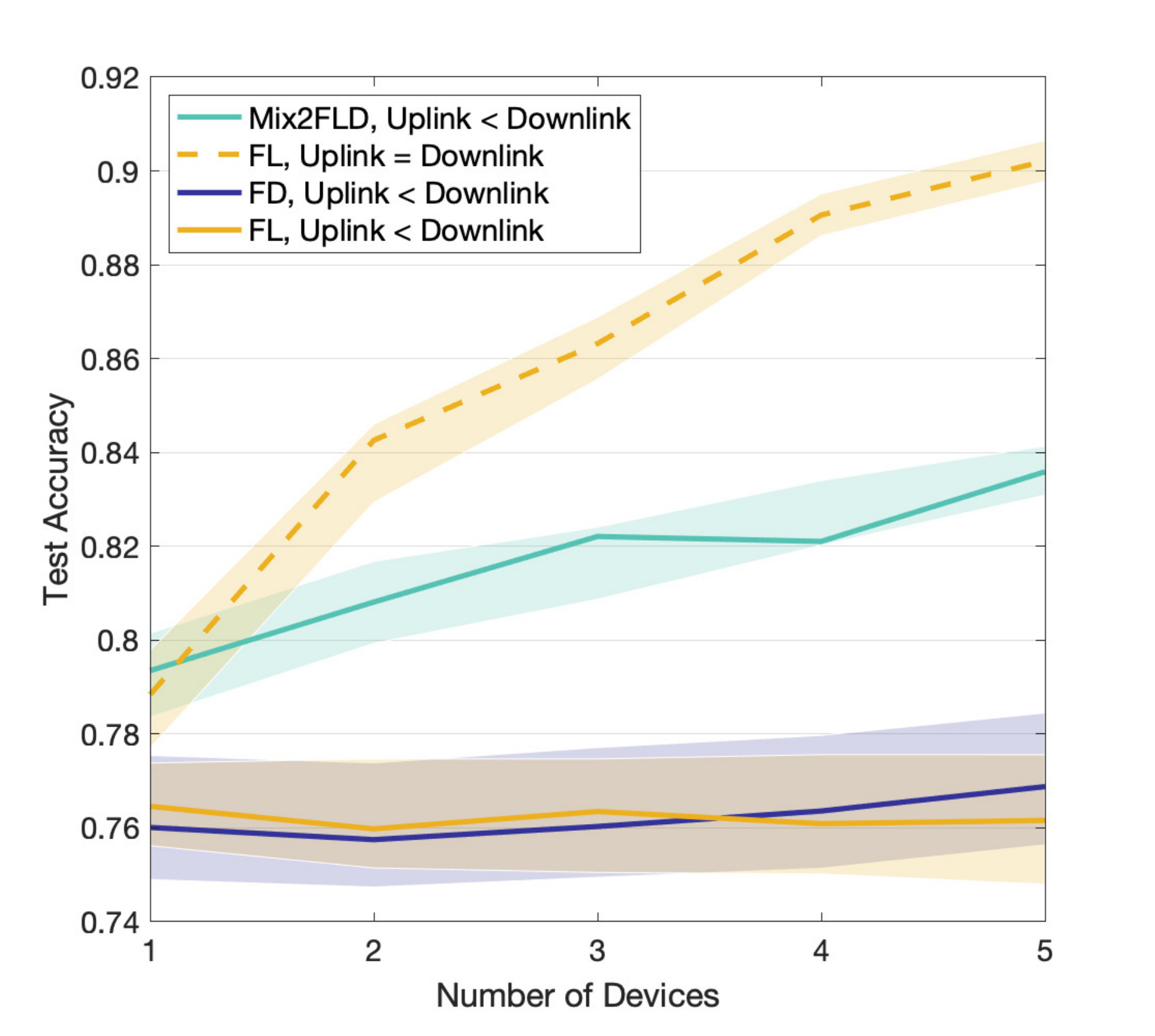}
    \caption{Test accuracy of Mix2FLD for a different number of workers, compared with FL and FD under symmetric (Uplink=Downlin) and asymmetric (Uplink < Downlink) channel capacities.}
    \label{fig:usecase_Mix2FLD}
\end{figure}

\subsection{Downlink FL After Uplink FD} \label{sec:mix2fl}



In mobile communication systems, uplink data rates are often much lower than downlink rates due to the limited transmission power of mobile devices \cite{JHParkTWC:15}. Therefore, FD (Sec. \ref{sec:method_FD}) is useful in the uplink thanks to its small payload sizes, whereas in the downlink FL (Sec. \ref{sec:method_FL}) is preferable in that exchanging model parameters commonly achieves higher accuracy than exchanging model outputs \cite{Oh20:CL}. 
To jointly exploit FD and FL under uplink-downlink asymmetric channels, we present an FL-after-FD algorithm combined with two-way Mixup (Mix2FLD). In Mix2FLD, the model outputs (i.e., logits) are uploaded to a server via FD, which should be converted into a global model whose parameters can be downloaded by and updated at each device using FL. Such a model output-to-parameter conversion is viable using KD (Sec. \ref{sec:principle_KD}) that updates the global model at the server by minimizing the difference between the uploaded outputs and the the outputs of the global model. This requires a handful of seed samples to generate the global model's outputs, which is a major challenge of its implementation due to the extra communication overhead and possible data privacy violation as highlighted in Sec. \ref{sec:challenges}.

In a classification task, we resolve the aforementioned problem by applying the Mixup method twice. Precisely, before uploading each device encodes multiple samples by them via Mixup (Sec. \ref{sec:principle_data_aug}). Then, the server decodes the Mixup-encoded samples uploaded from different devices by additionally superpositioning them, in a way that the decoded samples have one-hot labels. Such a decoding commonly improves accuracy particularly under non-IID data distributions \cite{Oh20:CL,shin2020xor}. Note that the encoding not only preserves raw data privacy but also reduces communication overhead (Sec. \ref{sec:challenges}) since the decoding based on the Mixup data augmentation can generate multiple synthetic seed samples by changing the superpositioning combinations.

Fig.~\ref{fig:usecase_Mix2FLD} first verifies our conjecture that FL achieves higher accuracy when the uplink channel capacity is as high as the downlink (Uplink = Downlink). However, when the uplink channel capacity is bottlenecked (Uplink $<$ Downlink), the accuracy of FL is significantly degraded due to its large payload sizes and the resultant frequent uploading failures within a target latency deadline. In this uplink-downlink asymmetric channel, Mix2FLD achieves higher accuracy with less variance than FL and FD.


\begin{figure}
  \centering
  \subfigure[A schematic illustration of XorMixFL.]{\includegraphics[width=\columnwidth]{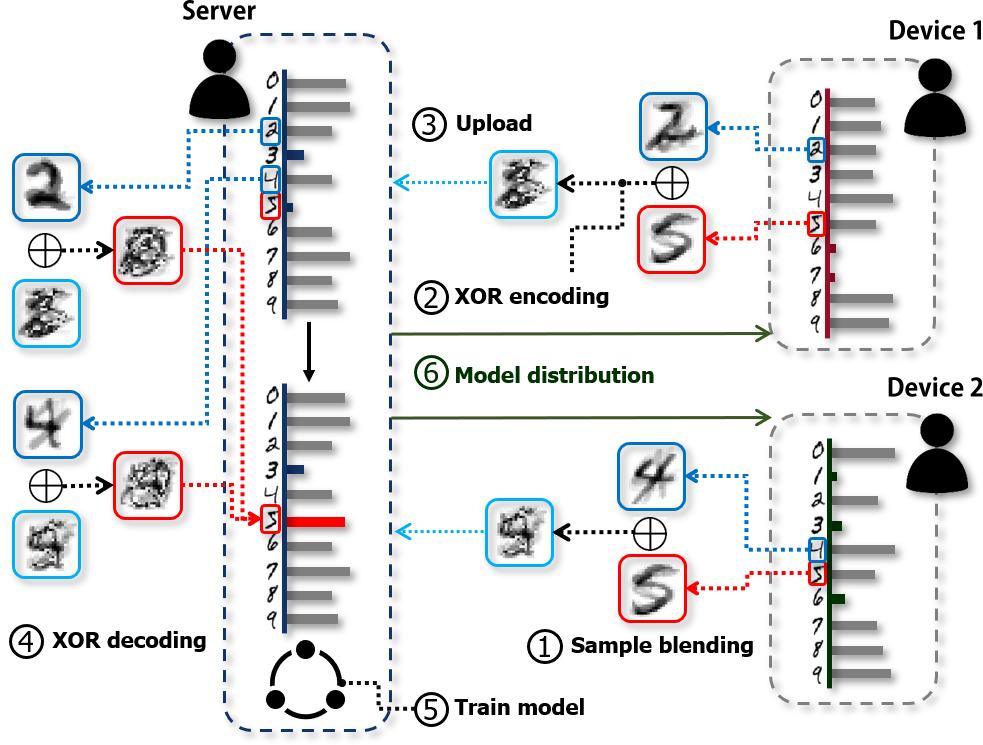}} 
  \subfigure[Test accuracy of XorMixFL.]{\includegraphics[width=.9\columnwidth]{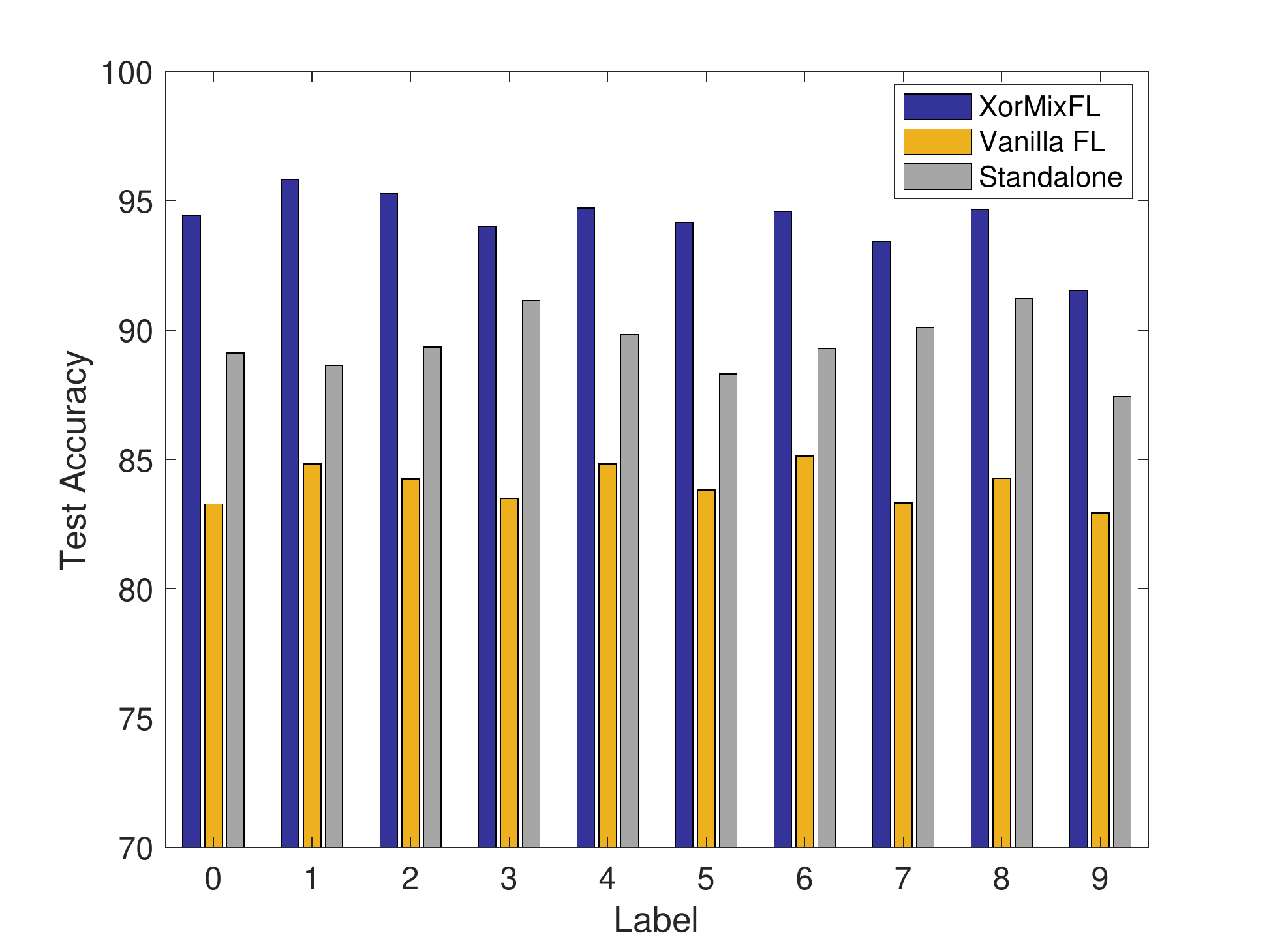}} 
  \caption{XorMixFL: (a) a schematic illustration in which non-IID data distributions are corrected by XorMixup data augmentation while preserving raw data privacy; and (b) test accuracy of XorMixFL for each label, compared to Vanilla FL and standalone traning.} \label{fig:usecase_XorMixFL}
  \vspace{-3mm}
\end{figure}

\subsection{One-Shot FL via XOR Mixup} \label{sec:xorfl}

Imbalanced data distributions could significantly degrade FL performance Sec. \ref{sec:method_FL})~\cite{park2018wireless,Zhao2018,Oh20:CL}. For the MNIST and CIFAR-10 datasets wherein each worker has scarce samples of specific labels, the classification accuracy is degraded by up to $11$\% and $51$\%, respectively, compared to the IID counterparts~\cite{zhao2018federated}. To correct such a non-IID data problem, a straightforward solution is to exchange and fill in missing raw samples, which may however violate data privacy. Alternatively, we apply an \emph{XOR based mixup data augmentation method (XorMixup)} that is extended to a novel one-shot FL framework, termed \emph{XorMixFL}. 

XorMixup was inspired by the Mixup data augmentation technique (Vanilla Mixup) producing a synthetic sample $(A+B)$ by linearly superpositioning two raw samples $A$ and $B$~(Sec. \ref{sec:principle_data_aug}) \cite{Zhang2018}. Similarly, XorMixup combines two samples not linearly but using the bit-wise XOR operation~$\oplus$ that has the following flipping property: $(A \oplus B) \oplus B = A$. To preserve the data privacy while generating realistic synthetic samples, (i) each worker encodes two local samples $(A \oplus B)$ that is exchanged with other devices, and (ii) the received $(A \oplus B)$ is decoded not using the original $B$ but a sample $B'$ stored in a different worker, which has the same label of $B$. Consequently, the decoding yields $(A \oplus B)\oplus B' = A'$ that reflects some key features of $A$ but is not the same as $A$. Owing to the mixing nature, both (i) and (ii) preserve raw data privacy across different workers, while (ii) improves the synthetic sample's authenticity, increasing one-shot FL accuracy as elaborated next.

As illustrated in Fig.~\ref{fig:usecase_XorMixFL}(a), by applying XorMixup to a one-shot FL framework having only one communication round~\cite{Guha:19,HybridFL}, each device in XorMixFL uploads its encoded seed samples to a server. The server decodes and augments the seed samples using its own base samples until all the samples are evenly distributed across labels. The server can be treated as one of the devices, or a parameter server storing an imbalanced dataset. Then, utilizing the reconstructed dataset, the server trains a global model that is downloaded by each device until convergence.
Under a non-IID MNIST dataset, simulation results in Fig.~\ref{fig:usecase_XorMixFL}(b) corroborate that XorMixFL achieves up to $8.13$\% and $17.6$\% higher accuracy than standalone ML and Vanilla FL, respectively.

\subsection{Tripartite SL for Medical Diagnosis} \label{sec:medicalsl}

In this use case, we study a privacy-preserving SL framework (Sec. \ref{sec:principle_model_split}) for multiple medical platforms (e.g., hospitals or e-health wearables). These platforms store their own privacy-sensitive medical data, and are willing to cooperatively train a global model by the aid of a server storing a fraction of the model. Specifically, we consider a medical image classification task, in which not only the raw samples (e.g., a chest X-ray images) but also their ground-truth labels (e.g., lung cancer diagnosis) are privacy sensitive. In an NN model, each raw sample is fed to the input layer, and its ground-truth label is compared with the model's prediction for loss calculation at the output layer. Therefore, to preserve the data privacy of each sample-and-label pair, both input and output layers should be stored by each platform, while the rest of the layers can be offloaded to the server, resulting in tripartite SL. This is in stark contrast to the standard bipartite SL where only the input layer is stored at each worker, while the remaining layers can be offloaded to the~server.

\begin{figure}[!t]
    \centering
    \subfigure[A schematic illustration of tripartite SL.]{\includegraphics[width=.9\columnwidth]{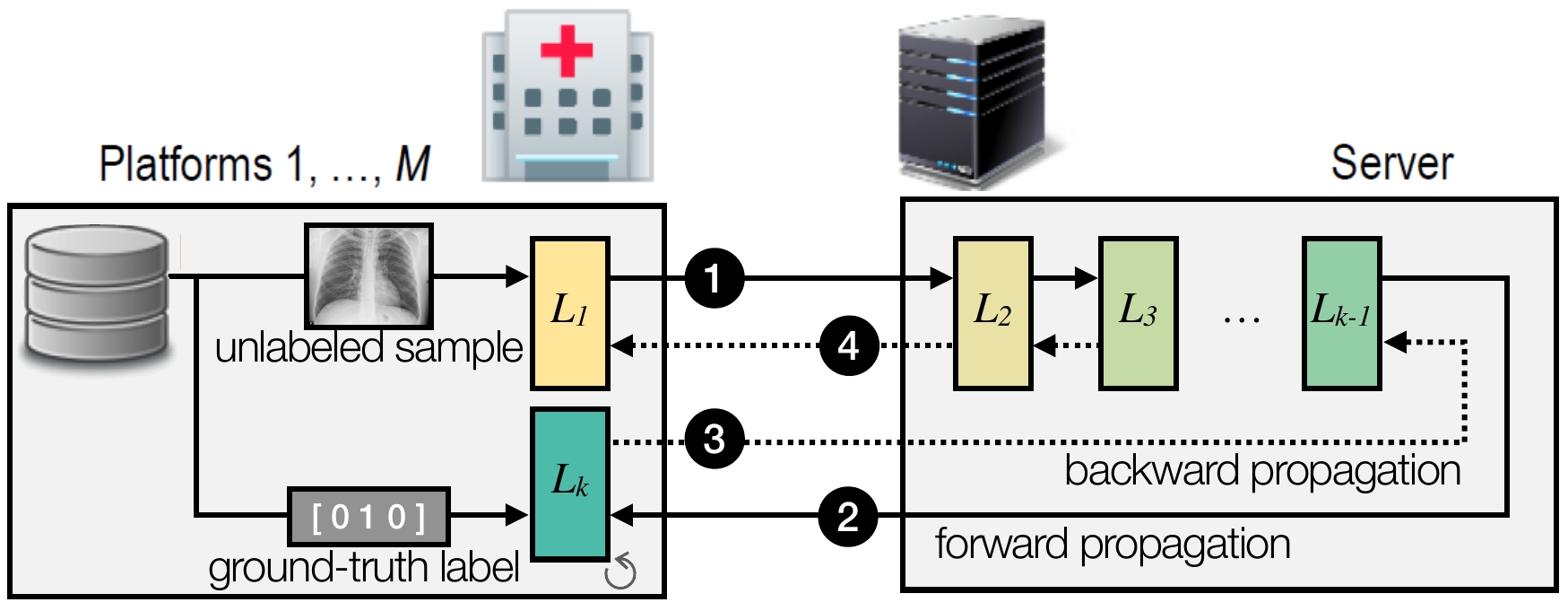}}\vspace{.5cm}
    \subfigure[A flowchart of tripartite SL.]{\includegraphics[width=.9\columnwidth]{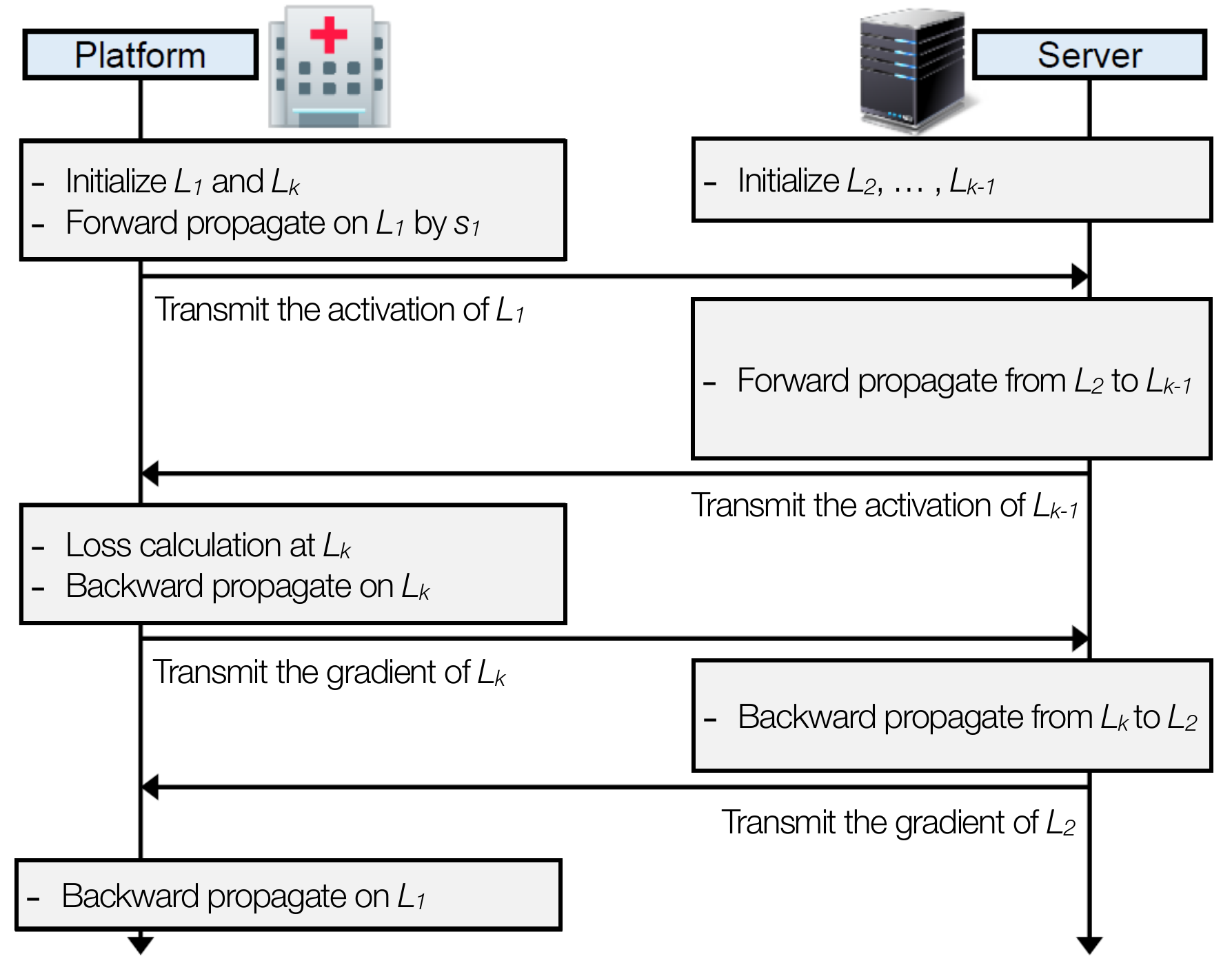}} \vspace{.5cm}
    \subfigure[The communication cost of tripartite SL until convergence under VGG and ResNet NN architectures, comapred to large-scale minibatch SGD.]{\includegraphics[trim=20 0 30 0, clip, width=\columnwidth]{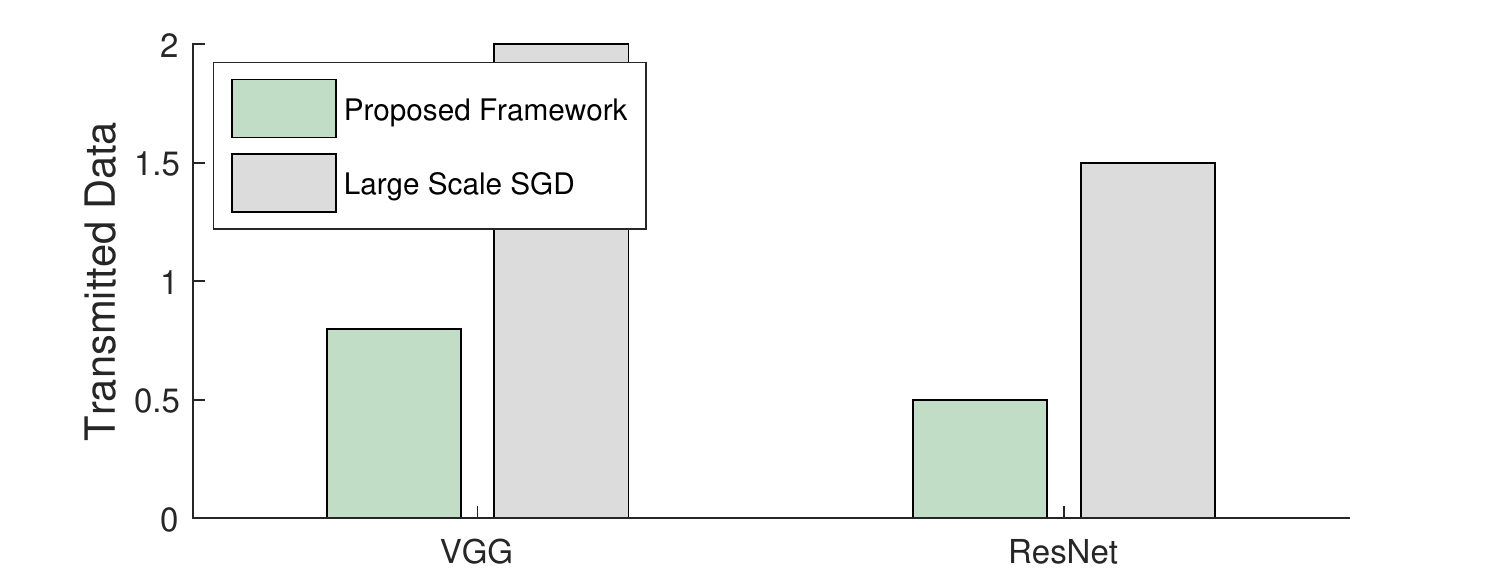}} \vspace{-3mm}
    \caption{Tripartite SL: (a) a schematic illustration in which the input and output layers are co-located with the data owner while the remaining hidden layers are stored at a server; (b) a flowchart clarifying forward and backward propagations; and (c) communication cost compared to the mini-batch~SGD.}\label{fig:usecase_MedicalSL}\vspace{-3mm}
    
\end{figure}

Following the aforementioned tripartite SL, as illustrated in Fig.~\ref{fig:usecase_MedicalSL}(a), we consider a single NN having $k$ layers whose intput layer $L_1$ and output layer $L_k$ are stored at each platform, while the rest is run at the server. As depicted in Fig.~\ref{fig:usecase_MedicalSL}(b), for each iteration, in the forward propagation, the activation of $L_1$ and $L_{k-1}$ are exchanged between a platform and the server without revealing raw samples. After calculating the loss a the output layer stored at the platform, the gradients of $L_k$ and $L_2$ are exchanged while hiding the ground-truth labels. While effective in preserving data privacy, the communication efficiency of tripartite SL is questionable due to frequent forward and backward propagations over wireless channels. 

To validate its communication efficiency, we compare tripartite SL with the large-scale minibatch stochastic gradient descent (LS-SGD)~\cite{SGD} by measuring their transmitted data until convergence under VGG and ResNet NN model architectures with a medical X-ray dataset, CheXpert~\cite{chest}. Fig.~\ref{fig:usecase_MedicalSL}(c) shows that under VGG, tripartite SL yields $0.8$\,GB transmitted data to achieve $95$\,\% test accuracy, while LS-SGD incurs $2$\,GB transmitted data with $55$\,\% accuracy. A similar tendency can be observed under ResNet, in which tripartite SL consumes $0.5$\,GB transmitted data with $75$\,\% accuracy, whereas LS-SGD results in $1.5$\,GB transmitted data with $10$\,\% accuracy. This experiment concludes that in spite of more frequent communications due to the layer splits and exchanging instantaneous forward/backward propagations, rather than periodically exchanging model parameters, tripartite SL ends up with achieving lower total communication cost until the convergence. This is viable thanks to its much less communication rounds (i.e., faster convergence) and much smaller communication payload~sizes.


\begin{figure}
    \centering

    \vspace{-2mm}
\end{figure}

    
    


\subsection{Channel and Packet Adaptive Parallel SL} \label{sec:parsl}



Vanilla SL is inefficient in terms of communication energy consumption when supporting multiple devices through wireless channels (Sec. \ref{sec:principle_model_split}).
Consider a server storing a common upper segment of NN layers that are associated with multiple devices storing its lower segments and feeding their own data samples. For these multiple devices, Vanilla SL is often implemented in a sequential manner preventing multiple devices to simultaneously connects with the server.
%
Concatenating the output features of multiple devices into a single large vector and feeding into the server can improve the SL performance \cite{Vepakomma:2018:Splita}, with increased transmission energy consumption in the uplink and back propagation overhead in the downlink.
%
Lastly, since the model structure cannot be dynamically adjusted during training and inference, for a fixed dimension of the server's input layer, the server needs to wait until the input layer is entirely filled with a predetermined number of devices or to pad arbitrary values for straggling devices due to poor channel conditions, increasing latency or degrading accuracy, respectively. 
Moreover, straggling devices due to intermittent connectivity under poor channel conditions (Sec. \ref{sec:challenges}) either increase waiting times of acquiring the input layer at the server or persuade the server to pad arbitrary values yielding loss of accuracy.
%

\begin{figure}[!t]
    \centering
    \subfigure[Performance comparison of Parallel SL in terms of test accuracy over training duration.]{\includegraphics[width=\columnwidth]{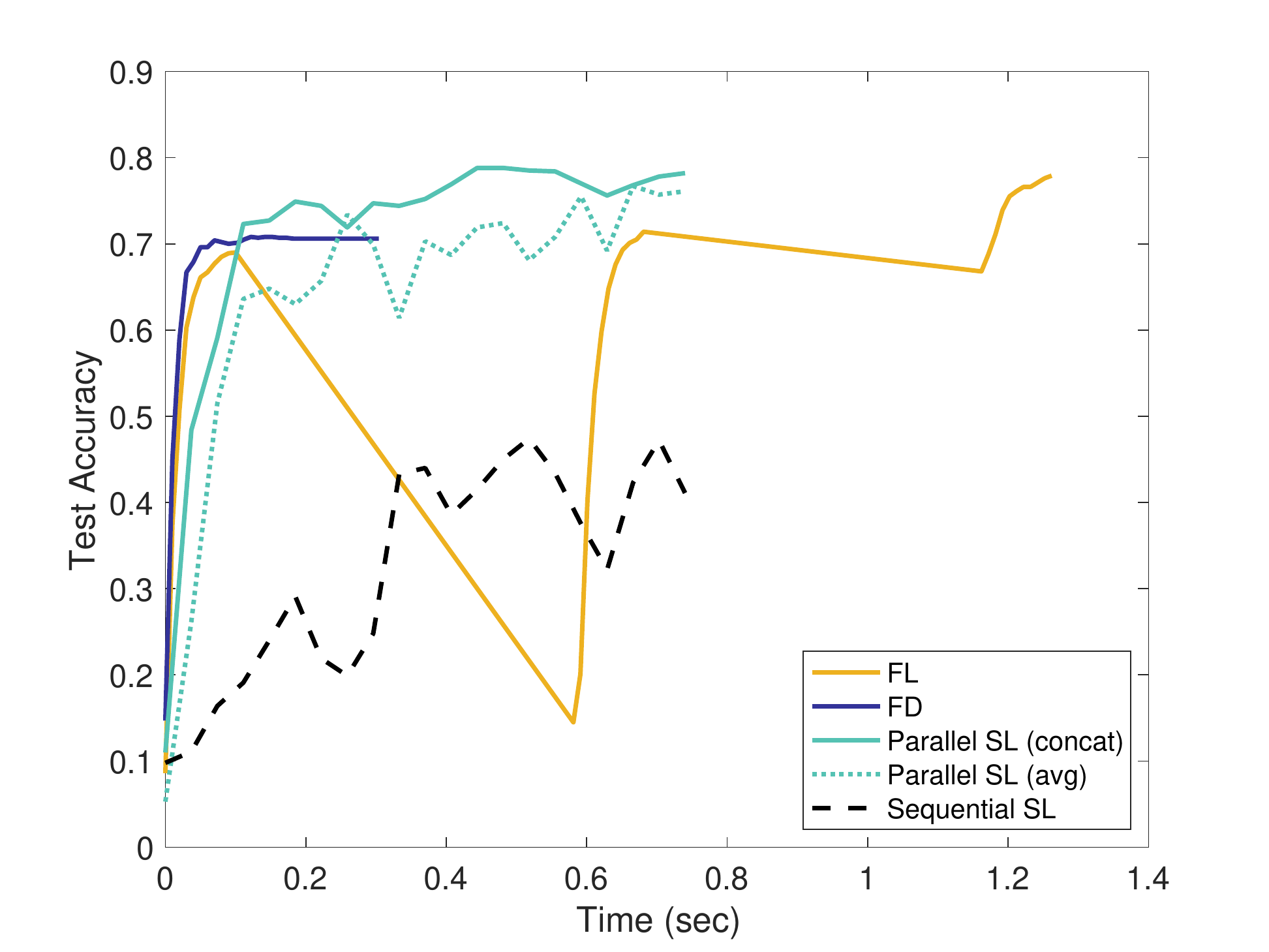}} 
    \subfigure[Impact of batch size on the tradeoff between test accuracy and training latency in Parallel SL.]{\includegraphics[width=\columnwidth]{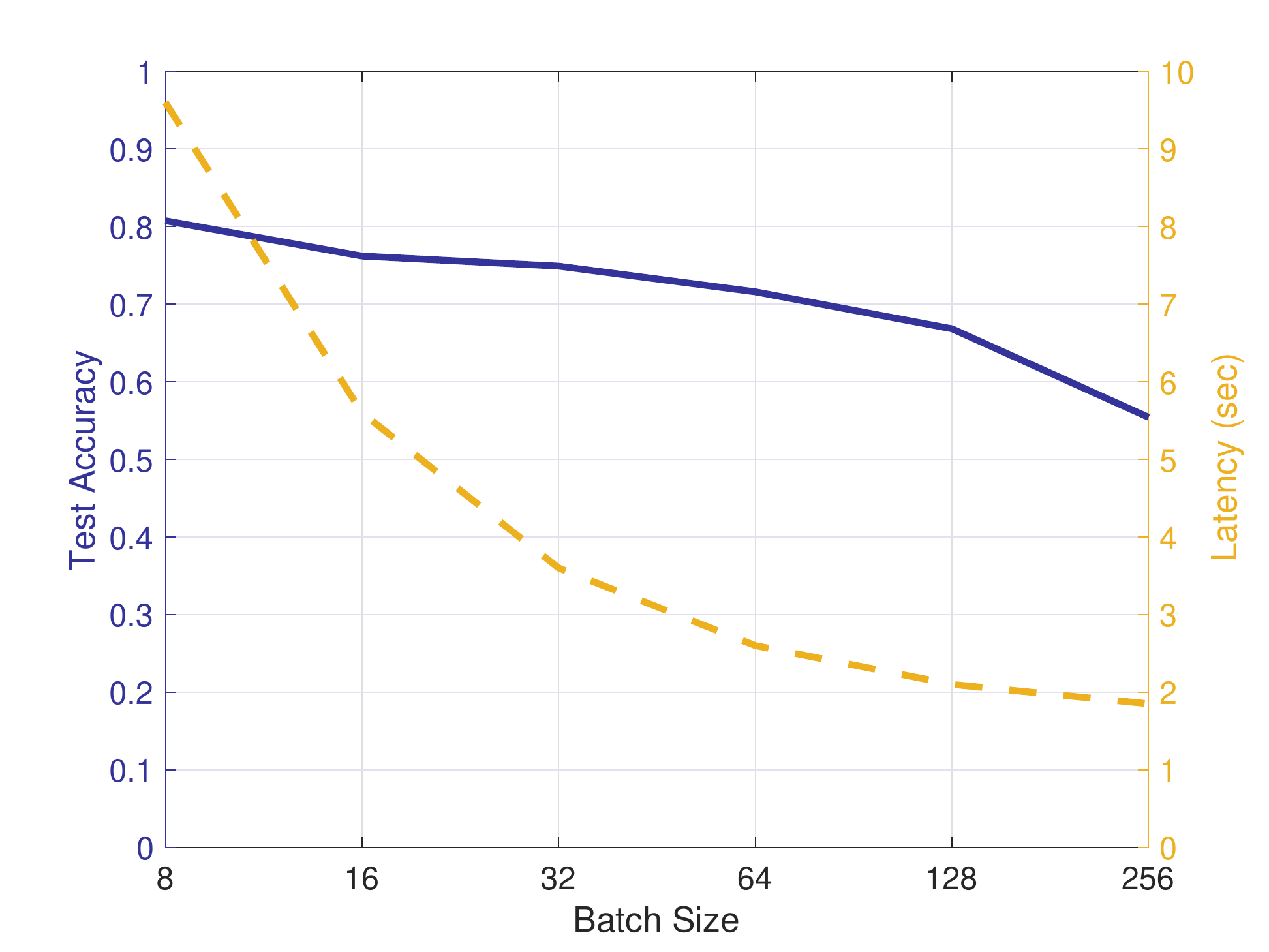}} 
    \caption{Parallel SL: (a) Learning curve of parallel SL compared to sequential (Vanilla) SL, FL, and FD ; and (b) accuracy (blue, left) vs. communication latency (yellow, right) for different batch sizes.} \label{fig:usecase_ParallelSL}
\end{figure}

%
In this view, parallel SL architecture utilizing feature averaging via Mixup augmentation-based (Section \ref{sec:principle_data_aug}) multiple devices' outputs super-positioning can be used \cite{verma2019manifold}.
%
Adopting feature averaging in contrast to output concatenation allows server's input dimension to remain fixed independent from the number of contributing devices, enabling communication and energy efficient scalability with low training latency as illustrated in Fig.~\ref{fig:usecase_ParallelSL}(a).
Additionally, controlling the batch size, packet sizes of the devices' cut layer's activation to be exchanged with the server can be controlled. 
With small batch sizes, the accuracy can be improved with the cost of degraded uplink data rates over short packets, which can be resolved by data aggregation as discussed in Sec.~\ref{sec:principle_short_packet}.
The tradeoff between test accuracy and training latency based on short packet aggregation for different choices of batch sizes is illustrated in Fig.~\ref{fig:usecase_ParallelSL}(b). 



\subsection{Heteromodal SL for mmWave Channel Prediction} \label{sec:heteromodalsl}



In this use case, we focus on predicting future mmWave channels by utilizing preceding mmWave signal received signal strength (RSS) history and image frames captured by two RGB depth (RGB-D) cameras mounted in different locations. Fusing these multiple modalities are essential in improving the prediction accuracy by complementing missing features one another. In particular, camera images involve useful features of blockage mobility patterns determining sudden line-of-sight (LOS) and NLOS transitions that are hardly observed from RSS, whereas RSS better describes short-term channel fluctuations for a given LOS or NLOS channel condition. Furthermore, the use of multiple cameras can overcome occlusions and missing frames (Sec. \ref{sec:challenges}) due to the limited field-of-views (FoVs) and insufficient frame rates of cameras, respectively. It is however challenging to fuse such multimodal and heterogeneous data. Indeed, these data are non-IID, under which FL and its variants cannot achieve high accuracy as highlighted in Sec. \ref{sec:method_FL}. 

\begin{figure}
	\centering
	\subfigure[Test RMSE.]{\includegraphics[width=0.49\columnwidth]{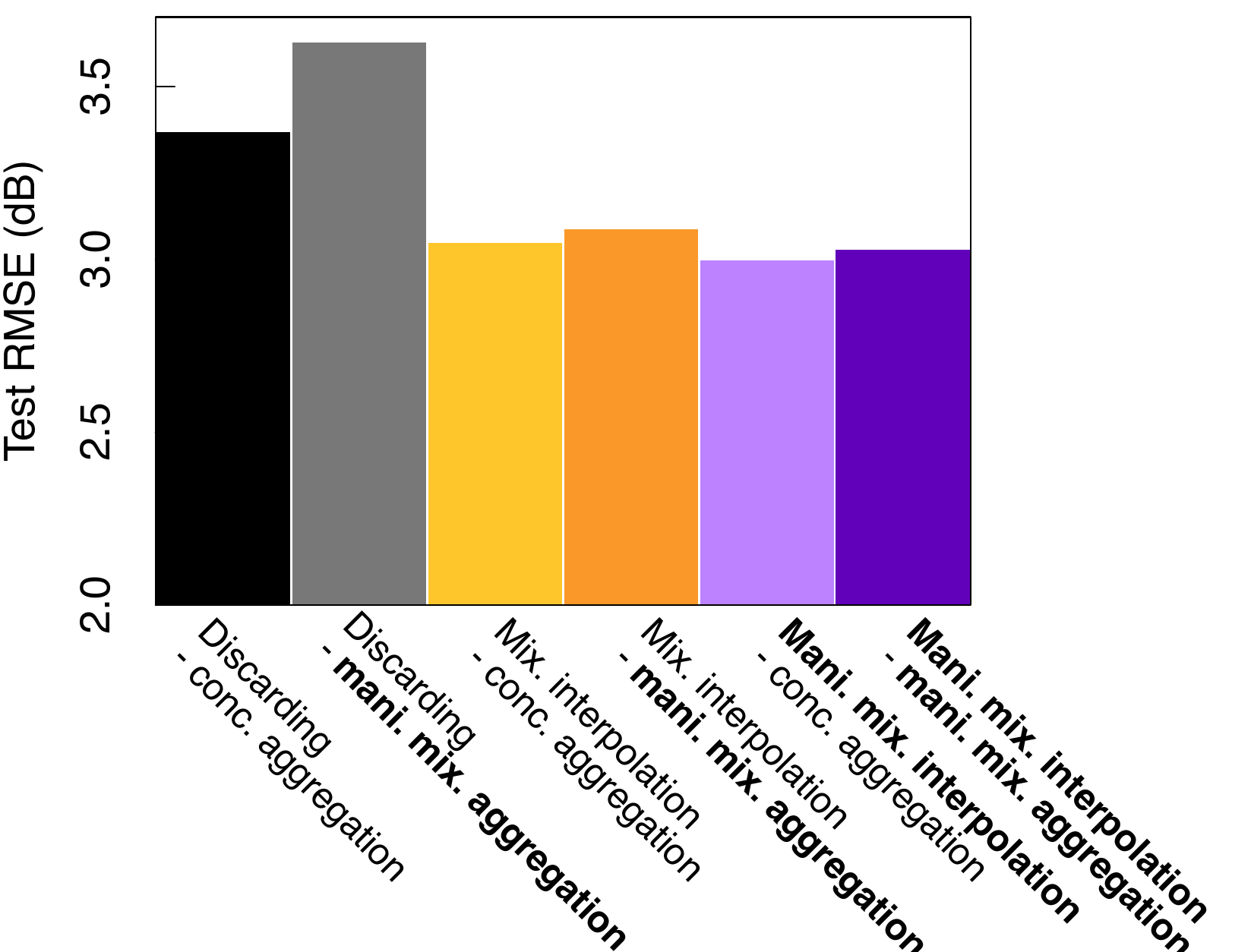}}
	\subfigure[Transmission latency for forward propagation signals.]{\includegraphics[width=0.49\columnwidth]{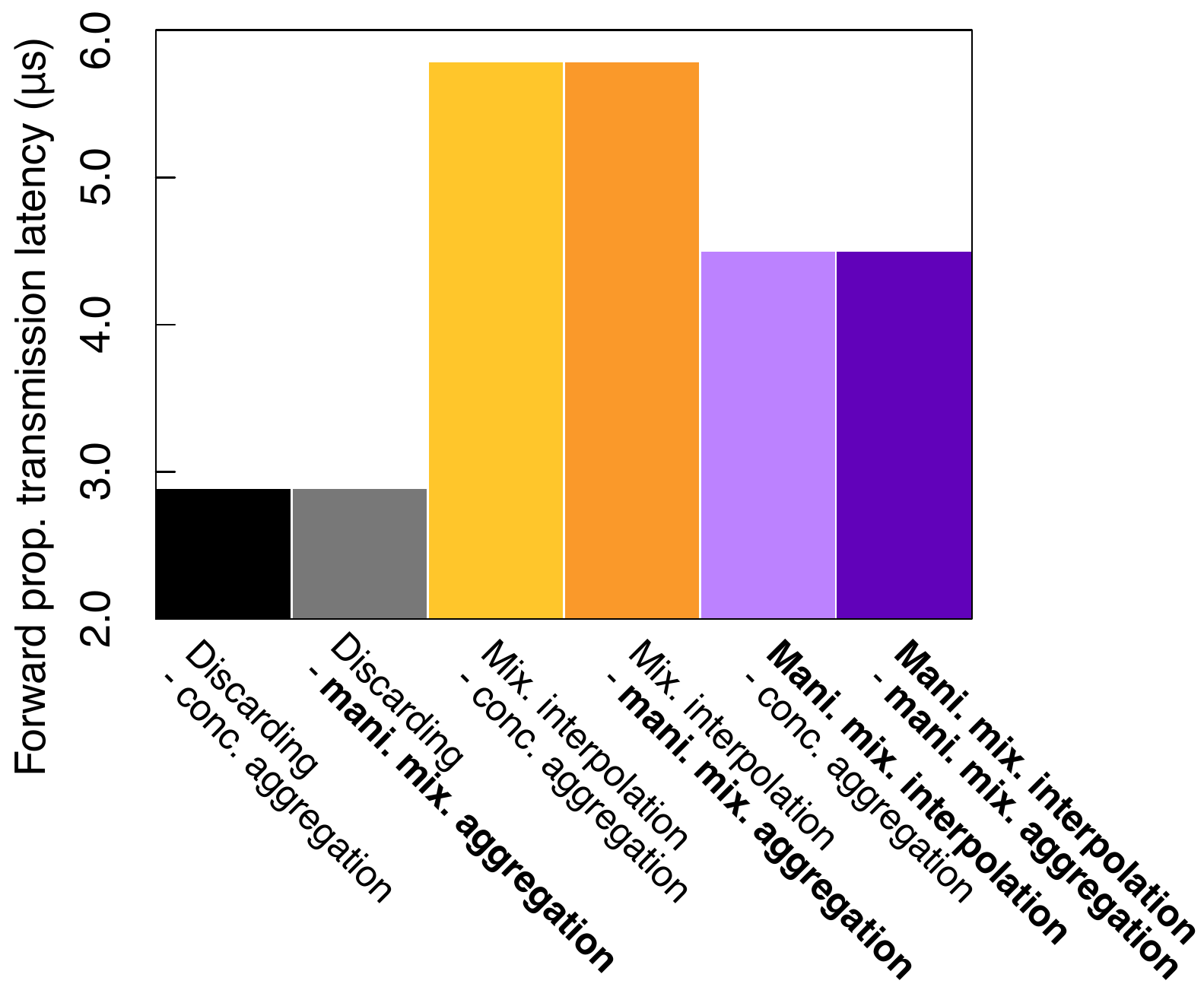}}
	\subfigure[Power consumption in cameras.]{\includegraphics[width=0.49\columnwidth]{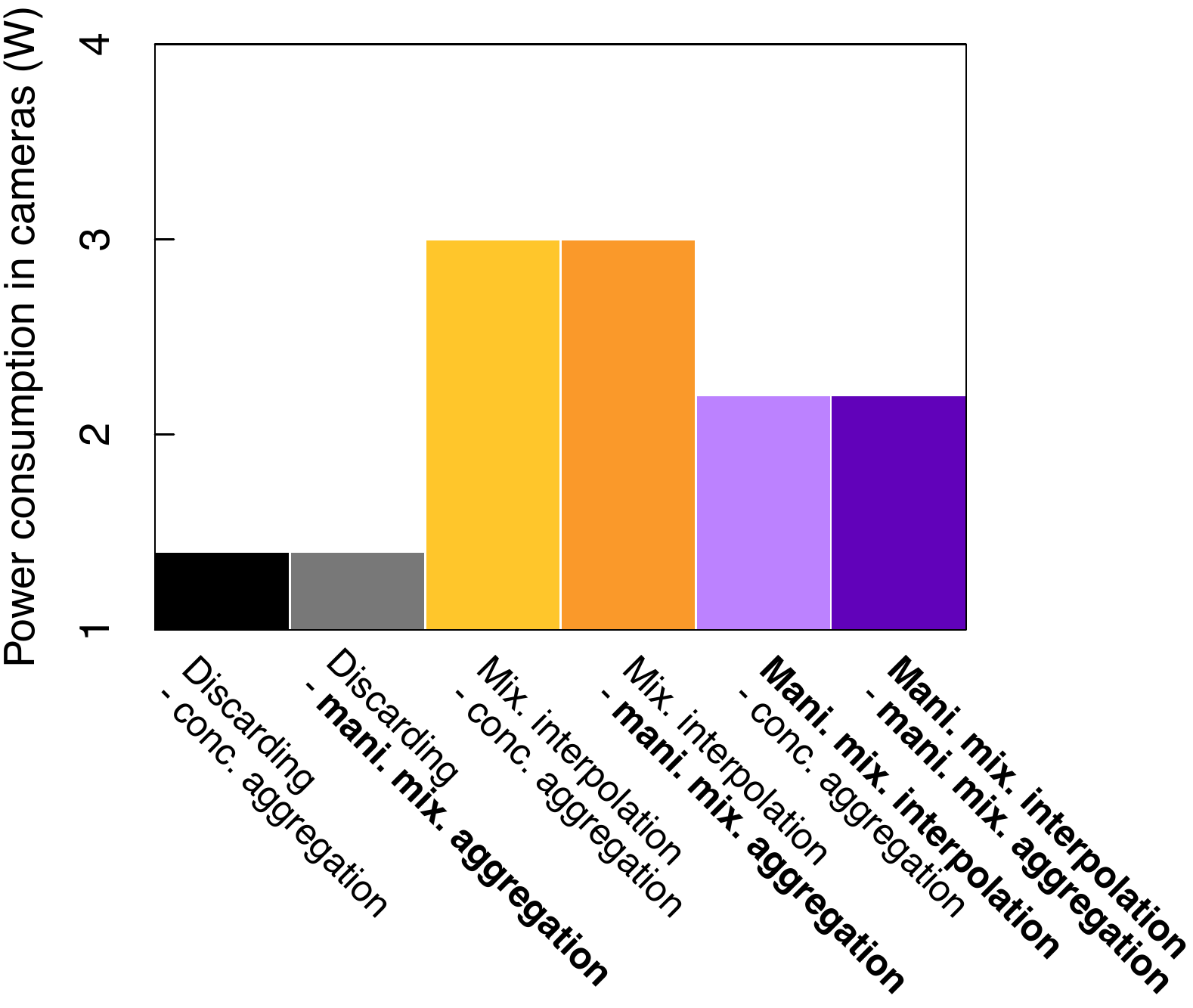}}
	\subfigure[Power consumption in BS.]{\includegraphics[width=0.49\columnwidth]{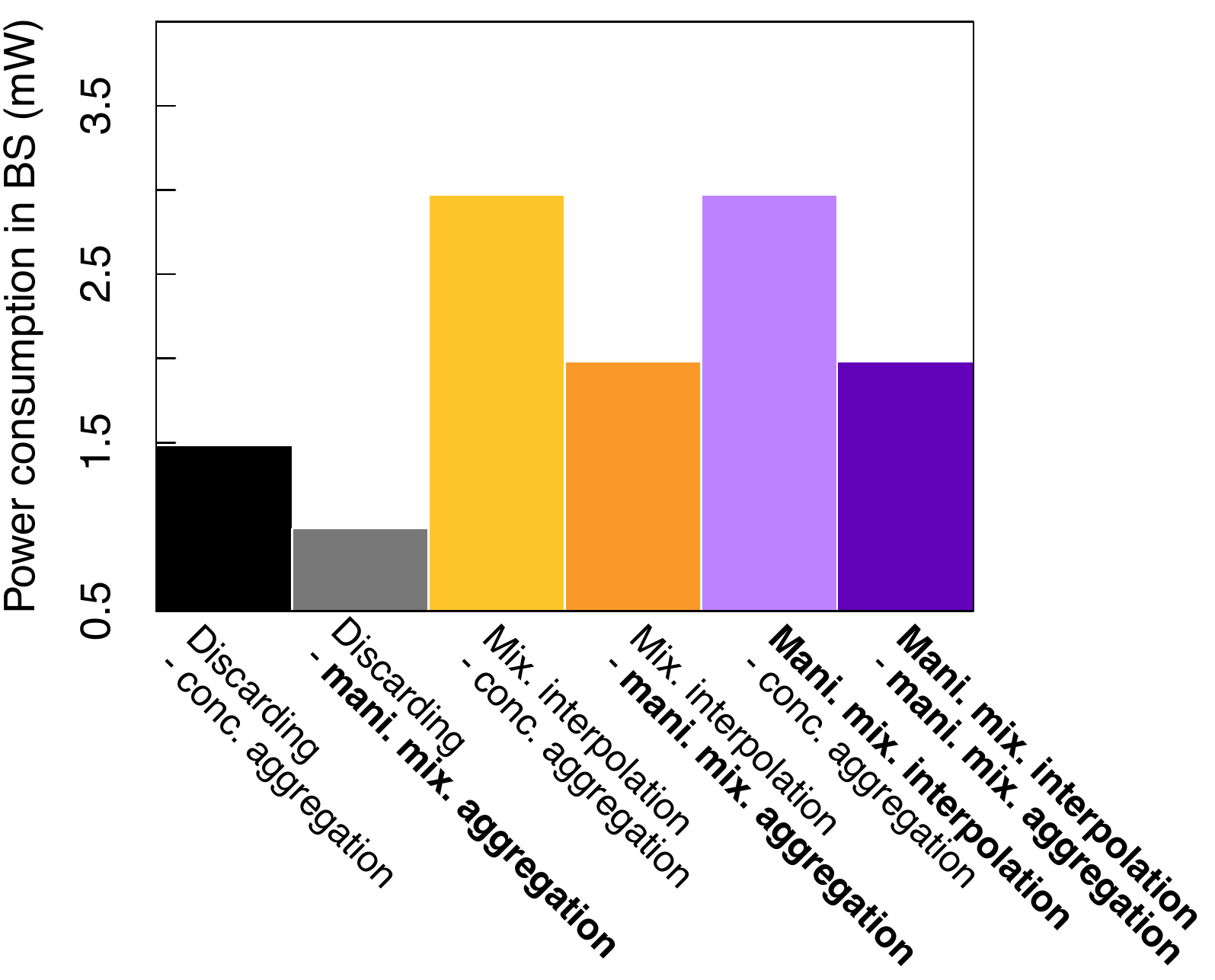}}\hspace{.5em}
	\caption{Performance comparison of Heteromodal SL in terms of: (a) accuracy, (b) communication latency, (c) energy consumption at cameras (workers), and (d) energy consumption at BS (server).}
	\label{fig:comprehensive_comparison}
\end{figure}

In this regard, a joint design of SL (Sec. \ref{sec:principle_model_split}) that is robust against non-IID data distributions \cite{Koda:GC20,koda2020distributed} and feature interpolation and averaging via the Mixup data augmentation (Sec. \ref{sec:principle_data_aug}) with heterogeneous FoVs and frame rates improving energy efficiency is considered.
%
%
In the SL design, each camera feeds a sequence of image frames into its convolutional and recurrent layers whose output is uploaded to a BS's fully connected layers at which the BS's uplink mmWave RSS is fused with the uploaded features from the cameras. 
%
%
The proposed SL framework is validated by simulation with data measured in a real experiment using 60\,GHz mmWave signals and two Kinect RGB-D cameras \cite{Kinect}. When predicting the future uplink mmWave RSS in 500\,ms by observing a sequence of RSS or image frames during 100\,ms.
%
%
To improve accuracy without degrading communication efficiency, the sequence of image features generated from the camera with a lower frame rate, missing feature elements are interpolated by equally superpositioning neighboring features via manifold Mixup (Sec.~\ref{sec:principle_data_aug}). 
Such an interpolation reduces the non-IIDness induced by the heterogeneous frame rates, yielding higher accuracy as shown in Fig. \ref{fig:comprehensive_comparison}(a). 
Note that this manifold Mixup for feature interpolation is performed within a sequence of features, whereas the aforementioned manifold Mixup for feature averaging is performed across the sequences uploaded from different cameras. 
Compared to a baseline scheme directly interpolating missing frames at cameras before transmissions, the aforementioned interpolation is performed at the BS after transmissions without increasing the communication payload sizes achieving low transmission latency as observed in~Fig. \ref{fig:comprehensive_comparison}(b), while yielding low power consumption at both cameras and BS as shown in Figs. \ref{fig:comprehensive_comparison}(c) and (d).









\section{ {Concluding Remarks}}\label{sec:conclusion}

Imbuing intelligence into edge devices enables low-latency and scalable decision-making at the network edge in 5G communication systems and beyond. On the other hand, updating outdated edge intelligence mandates communication with federating edge devices, improving the accuracy and reliability of the decision-making at the edge. To create greater synergy, this work has explored communication-efficient and distributed learning frameworks and their use cases by co-designing ML and communication principles under various challenges incurred by communication, computing, energy, and data privacy issues. The overarching goal of this article is to foster more fundamental research in this direction and bridge connections between communication and ML communities.

\vfill 
\pagebreak

\begin{IEEEbiography}
	[{\includegraphics[width=1in,height=1.25in,clip,keepaspectratio]{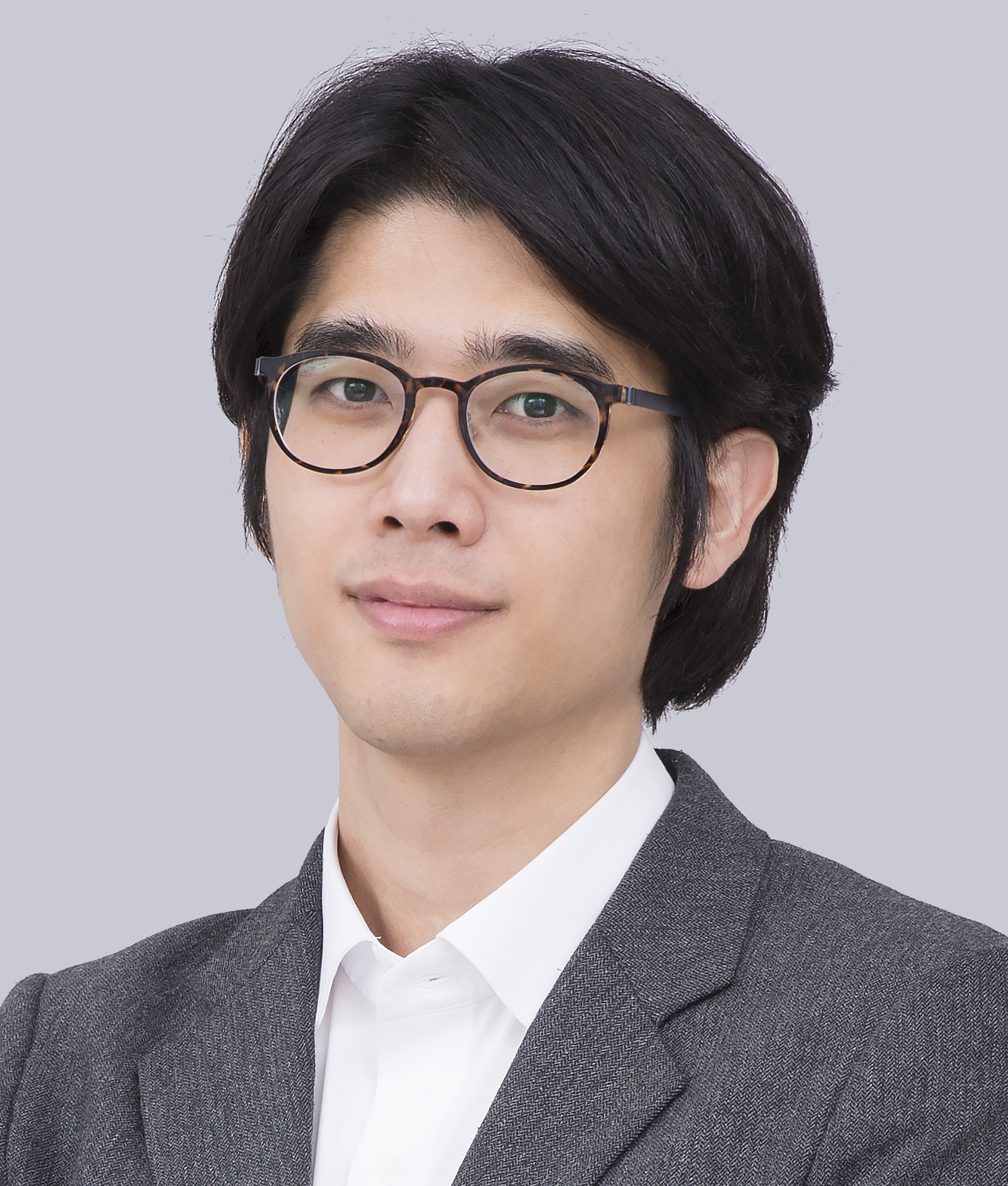}}]
	{Jihong Park} (S’09-M’16) is a Lecturer (assistant professor) at the School of IT, Deakin University, Australia. He received the B.S. and Ph.D. degrees from Yonsei University, Seoul, Korea, in 2009 and 2016, respectively. He was a Post-Doctoral Researcher with Aalborg University, Denmark, from 2016 to 2017; the University of Oulu, Finland, from 2018 to 2019. His recent research focus includes communication-efficient distributed machine learning, distributed control, and distributed ledger technology, as well as their applications for beyond 5G/6G communication systems. He served as a Conference/Workshop Program Committee Member for IEEE GLOBECOM, ICC, and WCNC, as well as NeurIPS, ICML, and IJCAI. He received the IEEE GLOBECOM Student Travel Grant in 2014, the IEEE Seoul Section Student Paper Contest Bronze Prize in 2014, and the 6th IDIS-ETNEWS (The Electronic Times) Paper Contest Award sponsored by the Ministry of Science, ICT, and Future Planning of Korea. Currently, he is an Associate Editor of Frontiers in Data Science for Communications, a Review Editor of Frontiers in Aerial and Space Networks, and a Guest Editor of MDPI Telecom SI on ``millimeter wave communiations and networking in 5G and beyond."
	
\end{IEEEbiography}

\begin{IEEEbiography}
	[{\includegraphics[width=1in,height=1.25in,clip,keepaspectratio]{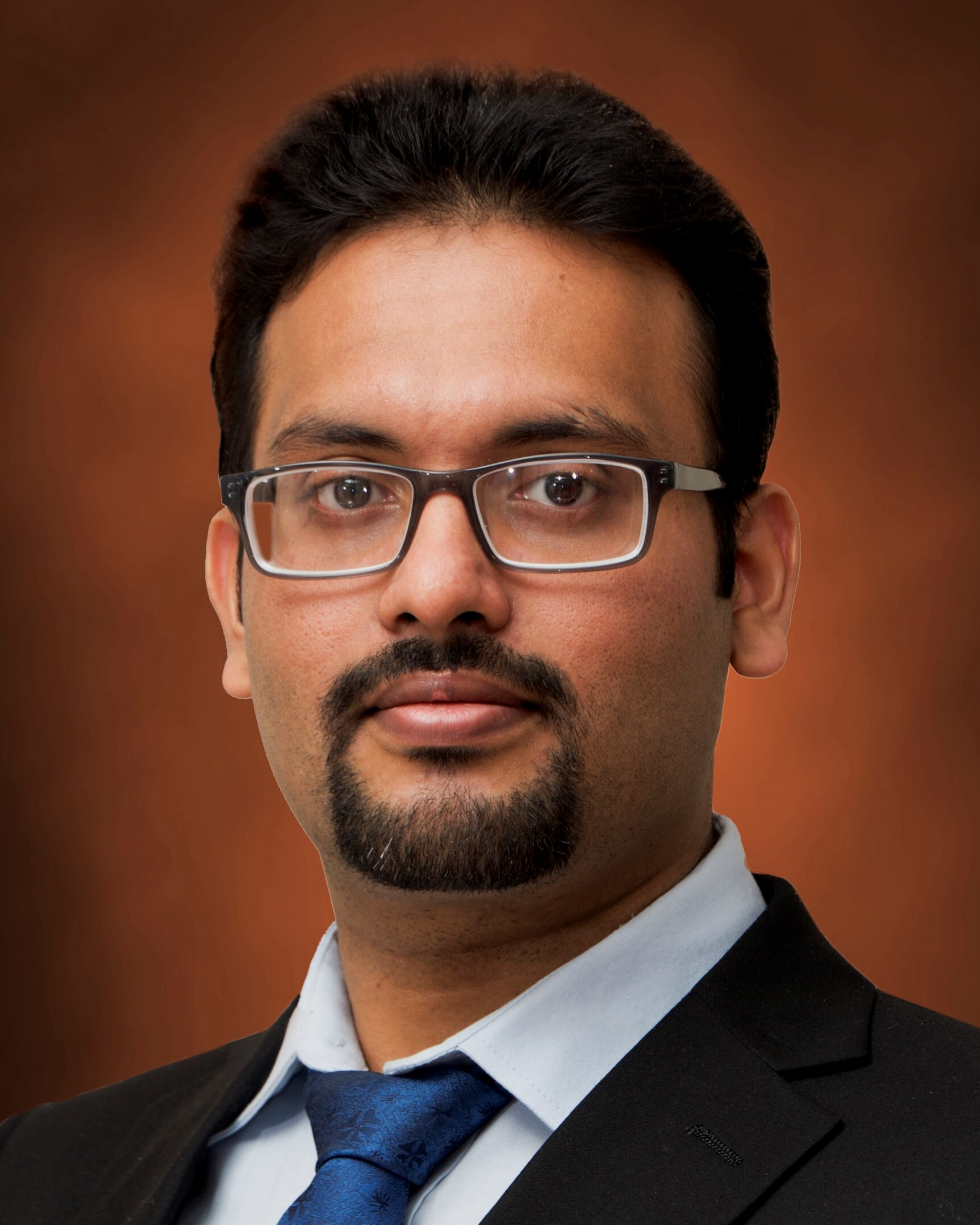}}]
	{Sumudu Samarakoon} (S'08-AM'18) received his B. Sc. Degree (Hons.) in Electronic and Telecommunication Engineering from the University of Moratuwa, Sri Lanka in 2009, the M. Eng. degree from the Asian Institute of Technology, Thailand in 2011, and Ph. D. degree in Communication Engineering from University of Oulu, Finland in 2017. He is currently working in Centre for Wireless Communications, University of Oulu, Finland as a post doctoral researcher. His main research interests are in heterogeneous networks, small cells, radio resource management, reinforcement learning, and game theory. In 2016, he received the Best Paper Award at the European Wireless Conference and Excellence Awards for innovators and the outstanding doctoral student in the Radio Technology Unit, CWC, University of Oulu.
\end{IEEEbiography}

\begin{IEEEbiography}
	 [{\includegraphics[width=1in,height=1.25in,clip,keepaspectratio]{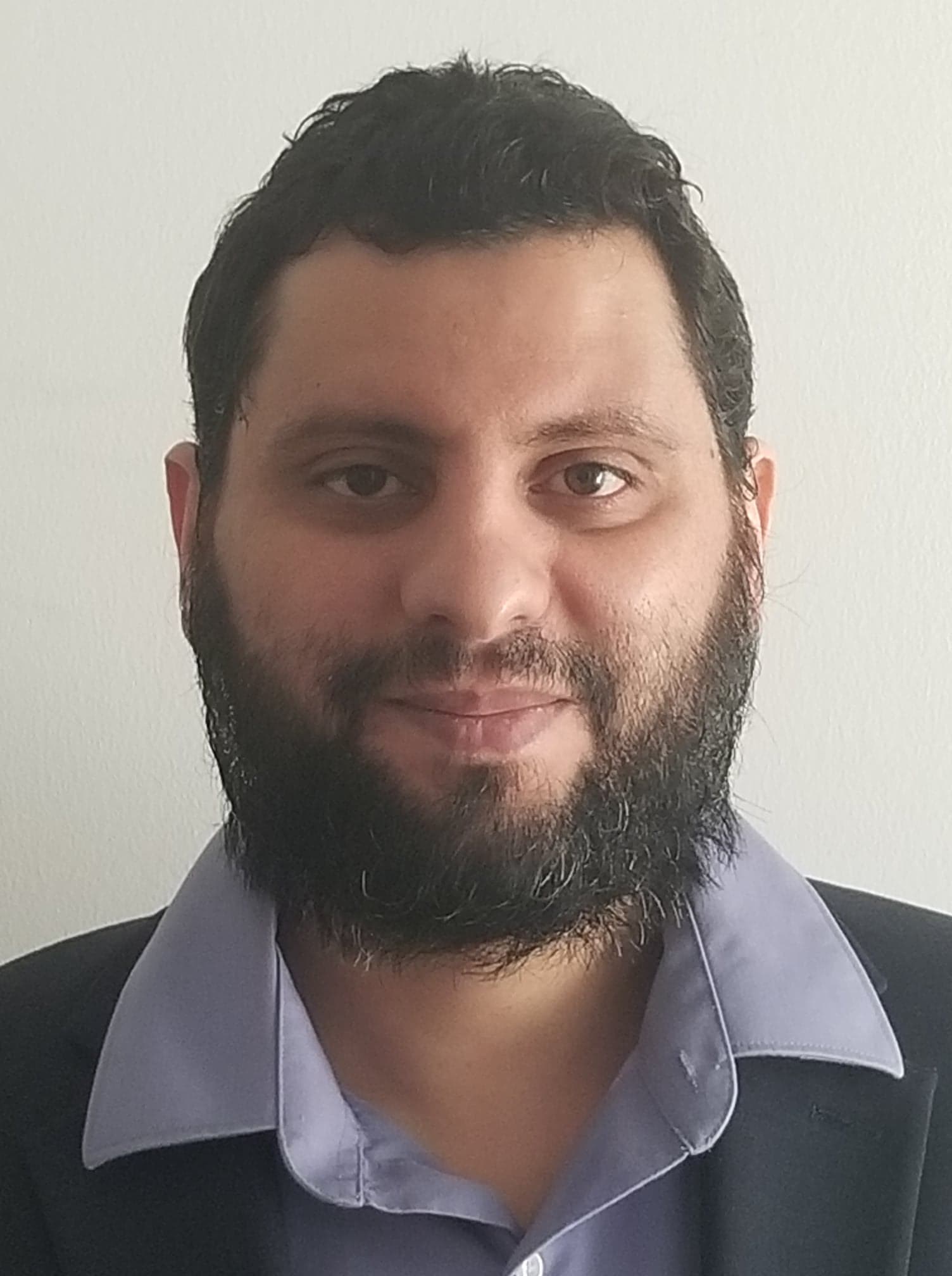}}]
	{Anis Elgabli} is a postdoctoral researcher at the Centre for Wireless Communications, University of Oulu. He received the B.Sc. degree in electrical and electronic engineering from the University of Tripoli, Libya, in 2004, the M.Eng. degree from UKM, Malaysia, in 2007, and MSc and PhD from the department of electrical and computer engineering, Purdue university, Indiana, USA  in 2015 and 2018 respectively. His main research interests are in heterogeneous networks, radio resource management, vehicular communication, video streaming, and distributed machine learning. He was the recipient of the best paper award in HotSpot workshop, 2018 (Infocom 2018).
\end{IEEEbiography}

\begin{IEEEbiography}
	[{\includegraphics[width=1in,height=1.25in,clip,keepaspectratio]{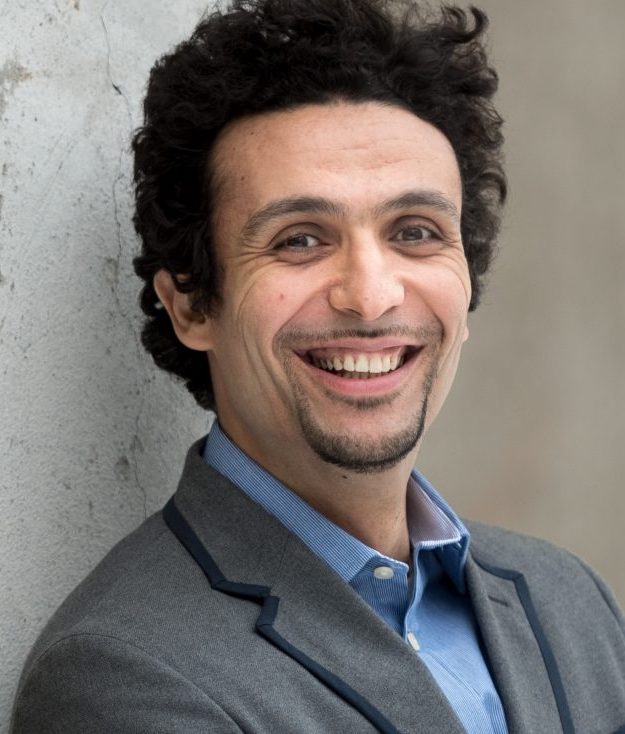}}]
	{Mehdi Bennis} is an Associate Professor at the Centre for Wireless Communications, University of Oulu, Finland, an Academy of Finland Research Fellow and head of the intelligent connectivity and networks/systems group (ICON). His main research interests are in radio resource management, heterogeneous networks, game theory and machine learning in 5G networks and beyond. He has co-authored one book and published more than 200 research papers in international conferences, journals and book chapters. He has been the recipient of several prestigious awards including the 2015 Fred W. Ellersick Prize from the IEEE Communications Society, the 2016 Best Tutorial Prize from the IEEE Communications Society, the 2017 EURASIP Best paper Award for the Journal of Wireless Communications and Networks, the all-University of Oulu award for research and the 2019 IEEE ComSoc Radio Communications Committee Early Achievement Award. Dr Bennis is an editor of IEEE TCOM.
\end{IEEEbiography}

\begin{IEEEbiography}
	[{\includegraphics[width=1in,height=1.25in,clip,keepaspectratio]{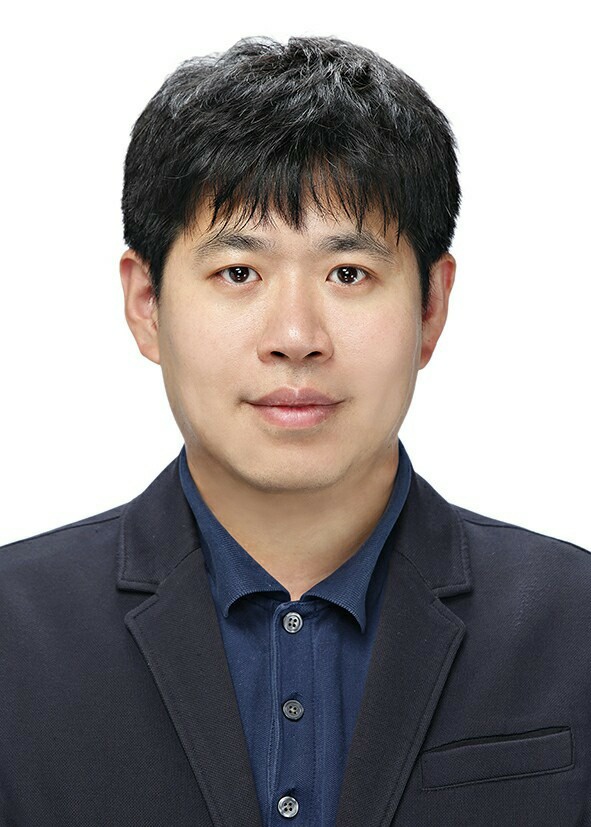}}]
	{Joongheon Kim} (M'06--SM'18) is currently an assistant professor of electrical engineering with Korea University, Seoul, Korea. He received his B.S. (2004) and M.S. (2006) in computer science and engineering from Korea University, Seoul, Korea; and his Ph.D. (2014) in computer science from the University of Southern California (USC), Los Angeles, CA, USA. Before joining Korea University as an assistant professor in 2019, he was with LG Electronics Seocho R\&D Campus as a research engineer (Seoul, Korea, 2006--2009), InterDigital as an intern (San Diego, CA, USA, 2012), Intel Corporation as a systems engineer (Santa Clara in Silicon Valley Area, CA, USA, 2013--2016), and Chung-Ang University as an assistant professor of computer science and engineering (Seoul, Korea, 2016--2019). He is a senior member of the IEEE. He was a recipient of the Annenberg Graduate Fellowship with his Ph.D. admission from USC (2009), Intel Corporation Next Generation and Standards (NGS) Division Recognition Award (2015), KICS Haedong Young Scholar Award (2018), IEEE Vehicular Technology Society (VTS) Seoul Chapter Award (2019), KICS Outstanding Contribution Award (2019), Gold Prize from IEEE Seoul Section Student Paper Contest (2019), and IEEE Systems Journal Best Paper Award (2020).
\end{IEEEbiography}

\begin{IEEEbiography}
	[{\includegraphics[width=1in,height=1.25in,clip,keepaspectratio]{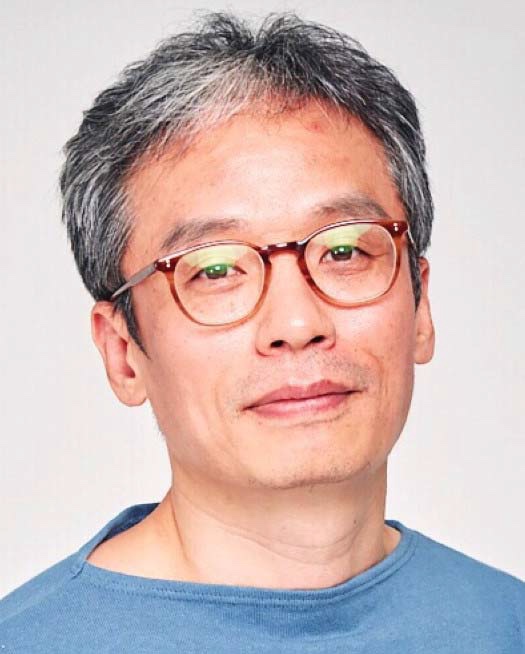}}]
	{Seong-Lyun Kim} is currently a Professor and Head of the School of Electrical \& Electronic Engineering, Yonsei University, Seoul, Korea, leading the Robotic \& Mobile Networks Laboratory (RAMO) and the Center for Flexible Radio (CFR+). He is co-directing H2020 EUK PriMO-5G project, and the chair of Smart Factory Committee of 5G Forum, Korea. He was an Assistant Professor of Radio Communication Systems at the Department of Signals, Sensors \& Systems, Royal Institute of Technology (KTH), Stockholm, Sweden. He was a Visiting Professor at the Control Engineering Group, Helsinki University of Technology (now Aalto), Finland, the KTH Center for Wireless Systems, and the Graduate School of Informatics, Kyoto University, Japan. He served as a technical committee member or a chair for various conferences, and an editorial board member of IEEE Transactions on Vehicular Technology, IEEE Communications Letters, Elsevier Control Engineering Practice, Elsevier ICT Express, and Journal of Communications and Network. His research interest includes radio resource management, information theory in wireless networks, collective intelligence, and robotic networks.
\end{IEEEbiography}

\begin{IEEEbiography}
	[{\includegraphics[width=1in,height=1.25in,clip,keepaspectratio]{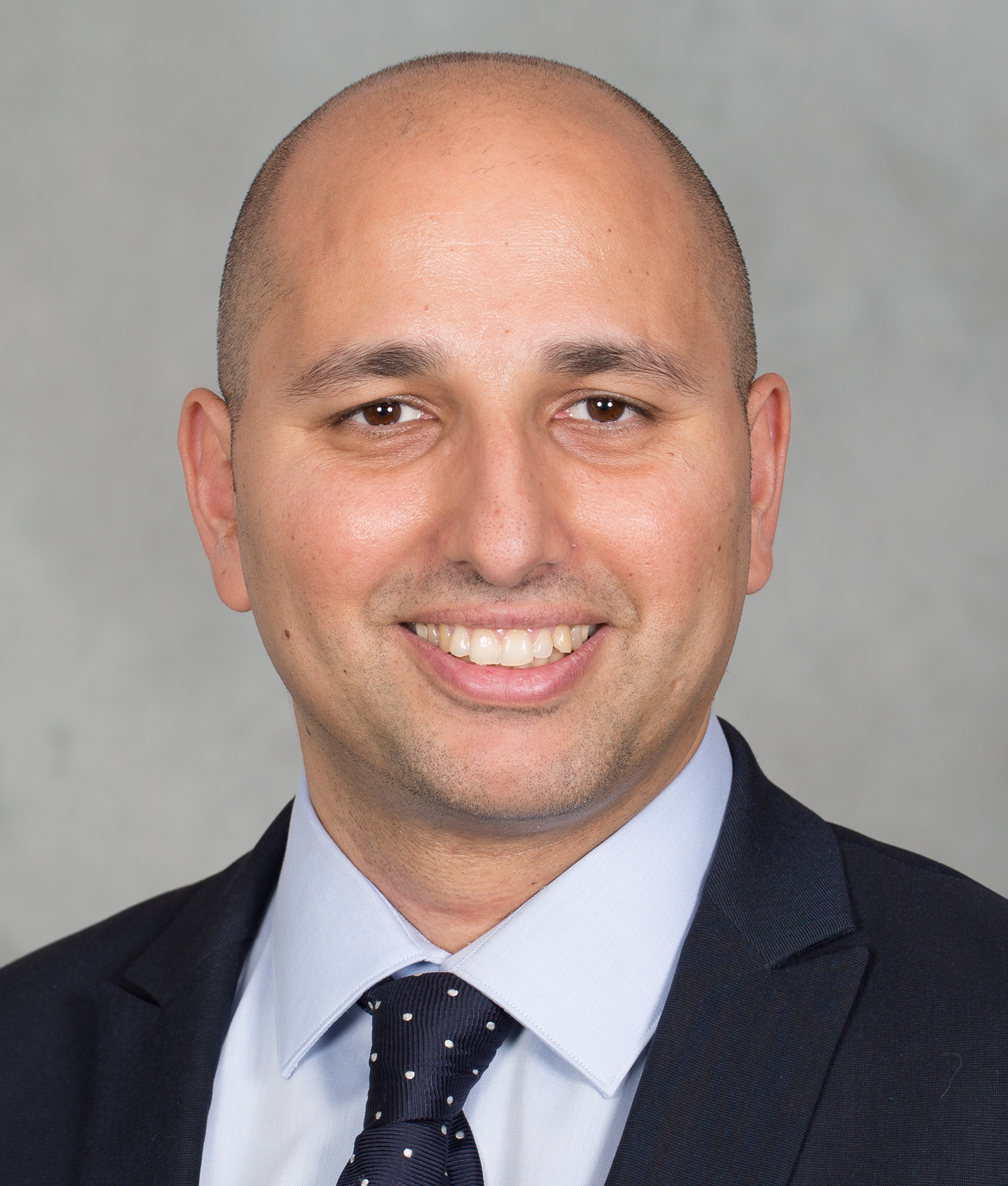}}]
	{M\'{e}rouane Debbah} (S'01-M'04-SM'08-F'15) received the M.Sc. and Ph.D. degrees from the Ecole Normale Supérieure Paris-Saclay, France. He was with Motorola Labs, Saclay, France, from 1999 to 2002, and also with the Vienna Research Center for Telecommunications, Vienna, Austria, until 2003. From 2003 to 2007, he was an Assistant Professor with the Mobile Communications Department, Institut Eurecom, Sophia Antipolis, France. From 2007 to 2014, he was the Director of the Alcatel-Lucent Chair on Flexible Radio. Since 2007, he has been a Full Professor with CentraleSupelec, Gif-sur-Yvette, France. Since 2014, he has been a Vice-President of the Huawei France Research Center and the Director of the Mathematical and Algorithmic Sciences Lab. He has managed 8 EU projects and more than 24 national and international projects. His research interests lie in fundamental mathematics, algorithms, statistics, information, and communication sciences research. He is an IEEE Fellow, a WWRF Fellow, and a Membre émérite SEE. He was a recipient of the ERC Grant MORE (Advanced Mathematical Tools for Complex Network Engineering) from 2012 to 2017. He was a recipient of the Mario Boella Award in 2005, the IEEE Glavieux Prize Award in 2011, and the Qualcomm Innovation Prize Award in 2012. He received 20 best paper awards, among which the 2007 IEEE GLOBECOM Best Paper Award, the Wi-Opt 2009 Best Paper Award, the 2010 Newcom++ Best Paper Award, the WUN CogCom Best Paper 2012 and 2013 Award, the 2014 WCNC Best Paper Award, the 2015 ICC Best Paper Award, the 2015 IEEE Communications Society Leonard G. Abraham Prize, the 2015 IEEE Communications Society Fred W. Ellersick Prize, the 2016 IEEE Communications Society Best Tutorial Paper Award, the 2016 European Wireless Best Paper Award, the 2017 Eurasip Best Paper Award, the 2018 IEEE Marconi Prize Paper Award, the 2019 IEEE Communications Society Young Author Best Paper Award and the Valuetools 2007, Valuetools 2008, CrownCom 2009, Valuetools 2012, SAM 2014, and 2017 IEEE Sweden VT-COM-IT Joint Chapter best student paper awards. He is an Associate Editor-in-Chief of the journal Random Matrix: Theory and Applications. He was an Associate Area Editor and Senior Area Editor of the IEEE TRANSACTIONS ON SIGNAL PROCESSING from 2011 to 2013 and from 2013 to 2014, respectively.
\end{IEEEbiography}

\bibliographystyle{ieeetr}  

\end{document}